\documentclass[10pt,twocolumn,letterpaper]{article}

\usepackage{cvpr}              

\usepackage{booktabs}
\usepackage{amssymb}
\usepackage{graphicx}
\usepackage{booktabs,pifont}
\newcommand{\cmark}{\ding{51}}  
\newcommand{\xmark}{\ding{55}}  
\usepackage[utf8]{inputenc}

\definecolor{cvprblue}{rgb}{0.21,0.49,0.74}
\usepackage[pagebackref,breaklinks,colorlinks,allcolors=cvprblue]{hyperref}

\def\paperID{xxxxx} 
\def\confName{CVPR}
\def\confYear{2026}

\title{ ReCCur: A Recursive Corner-Case Curation Framework for Robust Vision-Language Understanding in Open and Edge Scenarios}

\author{%
\makebox[\textwidth][c]{\fontsize{9}{10}\selectfont
Wei~Yihan\textsuperscript{1}, Shenghai~Yuan\textsuperscript{1}, Tianchen~Deng\textsuperscript{2}, Boyang~Lou\textsuperscript{3}, Enwen~Hu\textsuperscript{3}}\\[-0.2em]
\makebox[\textwidth][c]{\footnotesize
\textsuperscript{1}Nanyang Technological University \quad
\textsuperscript{2}Shanghai Jiao Tong University \quad
\textsuperscript{3}Beijing University of Posts and Telecommunications}\\[-0.2em]
\makebox[\textwidth][c]{\ttfamily\footnotesize
\{yihan005, shyuan\}@e.ntu.edu.sg, tianchen.deng@sjtu.edu.cn, \{woshiluobodan, owen.hu\}@bupt.edu.cn}%
}

\newcommand{\para}[1]{\noindent\textbf{#1.}\ }

\begin{document}
\maketitle
\begin{abstract}
\textbf{Corner cases}—rare or extreme scenarios driving real-world failures—are difficult to curate at scale: web data are noisy, labels brittle, and edge deployments preclude large retraining. We present \textbf{ReCCur} (\emph{Recursive Corner-Case Curation}), a low-compute, training-free-core framework that converts noisy web imagery into auditable, fine-grained labels via a multi-agent recursive pipeline. First, \emph{Large-Scale Data Acquisition \& Filtering} expands a domain vocabulary with a vision–language model (VLM), crawls the web, and enforces \emph{trimodal} (image/description/keyword) consistency with light human spot checks to yield refined candidates. Next, \emph{Mixture-of-Experts Knowledge Distillation} uses complementary encoders (e.g., CLIP/DINOv2/BEiT) for $k$NN voting with dual confidence activation and uncertainty sampling, converging to a high-precision set. Finally, \emph{Region-Evidence VLM Adversarial Labeling} pairs a Proposer (multi-granularity regions and semantic cues) with a Validator (global+local chained consistency) to produce explainable labels and close the loop. On realistic corner-case scenarios (e.g., flooded-car inspection), ReCCur runs on consumer-grade GPUs, steadily improves purity and separability, and requires minimal human supervision—providing a practical substrate for downstream training and evaluation under resource constraints. Our code and corner cases dataset will be made open-source upon acceptance.
\end{abstract}

\section{Introduction}

\textbf{Corner cases}---rare or extreme scenarios that break common training assumptions—are increasingly decisive for real-world reliability under open-world dynamics and shift. Models must stay robust with few or zero in-domain labels, while edge deployments restrict compute and preclude frequent large retraining. This motivates a central question: \emph{How can we continuously acquire, clean, and semantically enrich corner-case data at low cost with minimal human effort?}
\begin{figure*}[t]
    \centering
    \includegraphics[width=\textwidth]{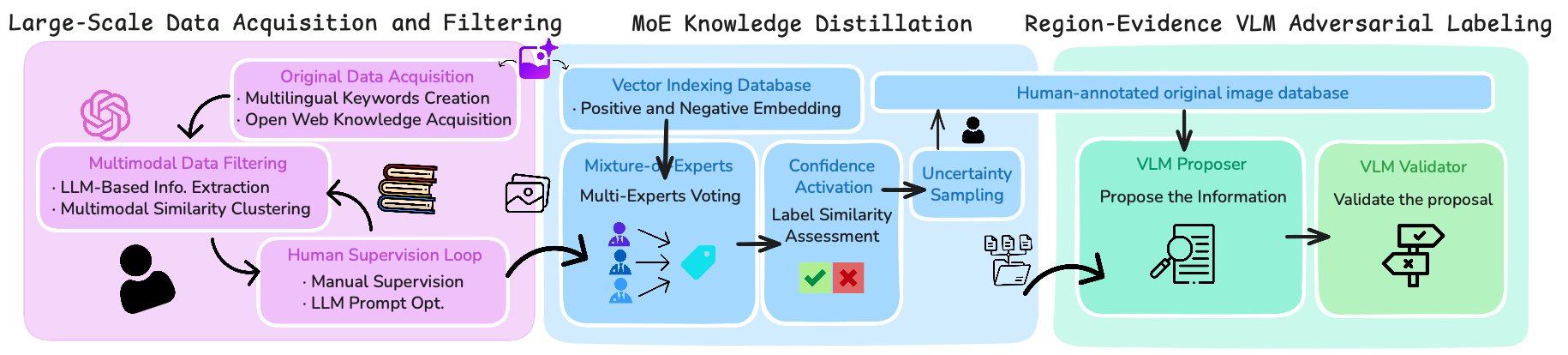}
    \caption{\textbf{Overview of ReCCur.} Multimodal acquisition/filtering, MoE labeling (with confidence + uncertainty), and region-evidence VLM progressively purify data and refine labels.}
    \label{fig:framework1}
\end{figure*}
Recent few-shot and open-set work trends toward three lines: (i) stabilizing decision boundaries in frozen/lightly-tuned spaces (margin optimization, local enhancement), (ii) augmenting with external semantics or retrieval (name expansion, transductive inference, retrieval-augmented evidence), and (iii) lightweight cross-domain adaptation. Effective as they are in curated settings, these methods typically \emph{require training or fine-tuning} and \emph{assume high-quality, controllable data/semantics}. Under noisy web acquisition, topic drift and label noise erode gains; large-scale filtering and re-annotation remain \emph{compute- and labor-heavy}, especially on edge hardware. Webly/self-supervised curation (confidence reweighting, clustering denoising, online/active selection) lowers cost but still relies on regular large-batch training dynamics and stable priors—fragile when corner cases dominate and resources are tight.

These limits expose three core challenges: \emph{Acquirability}—expand domain vocabularies and crawl the open web with high recall while controlling drift and long-tail noise; \emph{Trustworthiness}—filter massive candidates via multimodal consistency and within-class alignment \emph{without} large retraining, keeping rare positives while removing spurious ones; \emph{Deployability}—close the loop under low compute and tiny human budgets, producing explainable, fine-grained labels that are easy to audit and reuse.

We present \textbf{ReCCur} (\emph{Recursive Corner-Case Curation}), a training-free-core, multi-agent framework (Figure \ref{fig:framework1}) that runs a recursive pipeline—\emph{crawl $\rightarrow$ filter $\rightarrow$ distill $\rightarrow$ evidence-validated relabeling}—to turn noisy web imagery into high-confidence semantic labels. A vision–language model (VLM) expands vocabularies (multi-lingual), drives large-scale crawling, and measures trimodal consistency across image/description/keyword; clustering plus small-batch human spot checks partitions data and iteratively improves prompts, yielding a refined pool. A vector-indexed \emph{mixture-of-experts} (complementary visual encoders) performs $k$NN voting with topic/label confidence activation, while an uncertainty sampler surfaces low-alignment and boundary cases for minimal review, converging to a high-precision set. Finally, a region-evidence stage uses a \emph{Proposer} to generate multi-granularity regions and semantic cues, and a \emph{Validator} to run global+local chained-consistency checks, producing auditable, fine-grained labels that feed back into the loop.

\paragraph{Contributions.}
\begin{itemize}
    \item We introduce \textbf{ReCCur}, a recursive, training-free-core framework for corner-case data curation that achieves high-recall web acquisition and high-precision filtering under low compute and minimal human supervision.
    \item We combine trimodal consistency, mixture-of-experts vector voting with dual confidence activation, and uncertainty-driven human-in-the-loop to maximize data purity while tightly controlling annotation cost.
    \item We propose a region-evidence adversarial relabeling stage that yields explainable, fine-grained semantics and closes the curation loop for downstream reuse.
\end{itemize}

\section{Related Work}

\subsection{Corner cases and few-shot learning in vision}
Few-shot recognition in vision broadly follows three lines. \emph{(i) Feature-space stabilization} assumes fixed or lightly tuned backbones and reshapes embeddings so decision boundaries remain reliable with scarce labels—via instance-level max-margin objectives, operations directly in the frozen space to synthesize virtual support, attention-based regularization that emphasizes locally discriminative regions, and feature/semantic channel augmentations that expand intra-class coverage while preserving inter-class separability~\cite{article1,article2,article3,article4, article36, article37}. \emph{(ii) External semantics and retrieval} broaden coverage by expanding category names with LLMs, using transductive optimization over test batches, and aggregating neighborhoods from large retrieval corpora or stage-wise retrieval-augmented finetuning; open-world variants enrich CLIP with bidirectional image$\leftrightarrow$text alignment and prompt learning for detection and open-set recognition, improving robustness to unknowns~\cite{article5,article6,article7,article8,article9,article10,article11,article38}. \emph{(iii) Parameter-efficient adaptation} addresses domain shift with lightweight affine/orthogonal mappings or learnable attention modules that dynamically adjust feature distributions with minimal updates~\cite{article12,article13}. These directions markedly lift few-shot/open-set accuracy when semantics are transferable; however, they generally require (re)training and curated resources, and their gains can degrade under noisy, long-tail web acquisition and stringent edge-compute budgets—precisely the conditions where corner cases dominate.

\subsection{Data crawling and self-supervised dataset construction}
Large-scale web/self-supervised curation targets high recall with efficient denoising. Pipelines score image–text consistency, internalize reliability via self-consistency/confidence learning, reweight noisy tags, and apply clustering with balanced sampling to improve semantic purity without heavy manual labeling~\cite{article14,article16,article17}. To extend coverage and transferability, text-guided incremental expansion and external-memory retrieval leverage web-scale embeddings to enhance discrimination under substantial noise and label drift~\cite{article18,article19}. From an evaluation and scaling standpoint, unified candidate pools/protocols systematize filtering/resampling effects; dataset growth emphasizes \emph{quality$\times$diversity} with (re)captioning and multilingual augmentation; and scaling laws expose the coupling between filtering strength, dataset size, and available compute, arguing for budget-aware curation strategies~\cite{article20,article21,article2,article39}. Learnable/online/active pipelines further convert selection into trainable filters and use uncertainty, information gain, or relevance–specificity signals to surface high-value subsets under limited resources~\cite{article24,article25}. Despite progress, many approaches still presume regular large-batch training dynamics or stable priors; in edge scenarios with heavy noise and shifting topics, they leave a gap for training-free or minimally supervised governance of corner-case data.

\subsection{LLM agent and multi-agent voting}
Beyond simple majority voting, multi-agent LLM frameworks stress \emph{collaborative deliberation} and \emph{process control}. Round-table protocols run independent reasoning followed by mutual questioning to reach consensus~\cite{article26}. For long-context tasks, chain-of-agents decomposes problems with explicit handoffs, improving stability over extended reasoning windows~\cite{article28}; reflection and stepwise review add self-/cross-checks to reduce error accumulation~\cite{article29}.
A complementary line formalizes \emph{argumentation and adjudication}. Debate agents elicit divergent hypotheses and stress-test assumptions~\cite{article30,article44}; judge models score arguments to turn disagreements into supervision signals~\cite{article31,article32,article45,article46,article47}. Role-specialized systems (e.g., MetaGPT) import software-engineering abstractions—roles, interfaces, artifacts—to keep collaboration traceable and auditable~\cite{article33}. Applications beyond vision (financial decision-making; culturally sensitive alignment) highlight concept grounding and value balancing under heterogeneous priors~\cite{article34,article35}.
Guided by these principles, we orchestrate complementary experts to issue retrieval-based judgments with dual-confidence activation and \emph{uncertainty surfacing}, while region-aware VLM components supply localized, fine-grained evidence for pre-integration consistency checks. This coordination reconciles noisy signals, conserves human effort, and iteratively upgrades corner-case label quality under tight compute.

\section{Methodology}

\subsection{Large-Scale Data Acquisition and Filtering}

\begin{figure}[t]
    \centering
    \includegraphics[width=\linewidth]{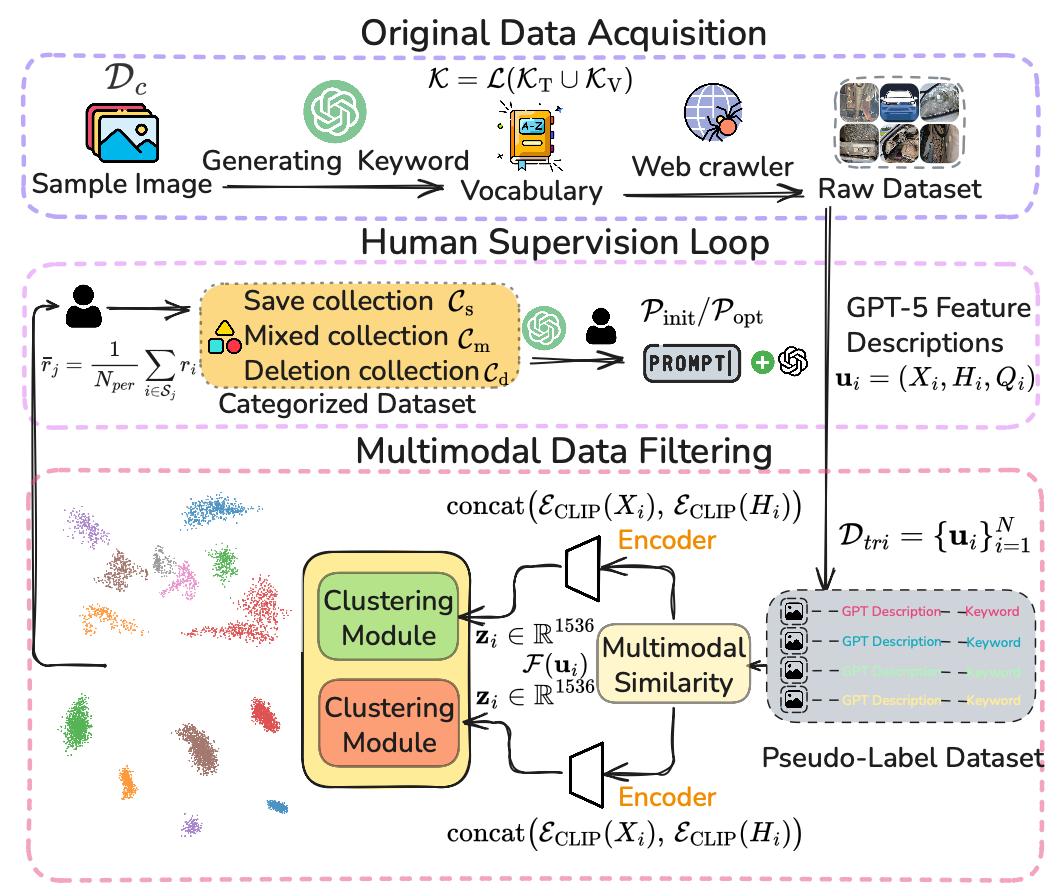}
    \caption{\textbf{Overall framework of Large-Scale Data Acquisition and Filtering.} LLM-expanded keywords guide crawling; the Human Supervision Loop refines prompts, adds descriptors, and categorizes samples into save/mixed/delete; vision–language clustering prunes noise to produce a pseudo-labeled, high-quality dataset.}
    \label{fig:framework}
    \vspace{-15pt}
\end{figure}

\para{Overall Structure} The overall pipeline in Figure~\ref{fig:framework} has three parts: data acquisition, data filtering, and human-in-the-loop supervision. Given $\mathcal{D}$, a VLM encodes images to feature descriptions $H_i$. For each sample, the triplet $\mathbf{u}_i=(X_i,H_i,Q_i)$ enables multimodal similarity for coarse filtering. A clustering module groups the enhanced embeddings $\mathbf{z}_i$, after which the clusters are assigned to three classes using a small labeled set. Guided by these annotations, the VLM is re-prompted with optimized instructions to produce the refined dataset $\Pi$.

\vspace{0.5em}

\para{Original Data Acquisition} Given a small set of sample images, each category $c$ is represented by
$\mathcal{D}_c=\{X_{c,1}, X_{c,2},\dots, X_{c,N_c}\}$, where $X_{c,i}$ denotes the $i$-th image of category $c$ and $N_c$ is the number of samples. Each category is specified by a visual prompt together with a textual prompt that encodes the dataset context and task objective.

Using a vision–language model, this information is expanded into a set of highly relevant crawler keyword descriptions
$\mathcal{K}_{\mathrm{V}} = \mathcal{G}_{\mathrm{V}}(\mathcal{D}_c)$, suitable for large-scale web collection.

To maximize data acquisition, we adopt a dual strategy: in addition to generating crawler keywords from joint vision–language prompts, we also employ textual prompts alone to obtain
$\mathcal{K}_{\mathrm{T}} = \mathcal{G}_{\mathrm{T}}(\mathcal{D}_c)$, thereby increasing the volume of collected data.
To further enhance crawler diversity, we apply a multilingual lexicon $\mathcal{L}(\cdot)$ for keyword augmentation, yielding the final keyword corpus

\[
\mathcal{K} = \mathcal{L}(\mathcal{K}_{\mathrm{T}} \cup \mathcal{K}_{\mathrm{V}}).
\]

Based on this augmented corpus, we perform keyword-guided web crawling to construct a prospective large-scale dataset $\mathcal{D}$ from open web sources.





\vspace{0.5em}
\para{Multimodal Data Filtering} Open-web data is noisy: despite retrieving $\mathcal{D}$ with high-relevance keywords $Q_i$, many images remain irrelevant or weakly related, so we filter them. Using prompts $\mathcal{P}_{\mathrm{init}}$ or $\mathcal{P}_{\mathrm{opt}}$ with ChatGPT-5, we generate feature descriptions $H_i$ for each $X_i\in\mathcal{D}$, forming image–feature–keyword triplets
\[
\mathbf{u}_i = (X_i, H_i, Q_i), \quad \mathcal{D}_{\mathrm{tri}} = \{\mathbf{u}_i\}_{i=1}^N.
\]

With these tuples, we embed samples using pre-trained CLIP and compute cosine similarities between image and description, between image and keyword, and between description and keyword, producing three complementary metrics. Modality-specific weights then fuse them into a single multimodal similarity score

\[ \mathrm{Sim}_{\mathrm{mm}}(\mathbf{u}_i) = \mathcal{F}(\mathbf{u}_i). \]

Samples with multimodal similarity below $\tau_{\mathrm{mm}}$ go to a low-similarity set. Those above $\tau_{\mathrm{mm}}$ stay in a high-similarity set for clustering/filtering, removing most irrelevant or weakly related images and yielding $\Pi$. Finally, each sample is represented by a vector obtained by concatenating $H_i$ and the embedding of image $X_i$ from a pre-trained CLIP encoder, yielding a semantically enhanced embedding

\[ \mathbf{z}_i = \mathrm{concat}\big(\mathcal{E}_{\mathrm{CLIP}}(X_i),\, \mathcal{E}_{\mathrm{CLIP}}(H_i)\big) \in \mathbb{R}^{1536}. \]


By clustering the feature vectors $\mathbf{z}_i$ of all samples in $\Pi$, we obtain

\[ \mathcal{C} = \{\, \mathcal{C}_1, \mathcal{C}_2, \dots, \mathcal{C}_J \,\}. \]

\para{Human Supervision Loop} 
After clustering, we obtain $\mathcal{C}$, each $\mathcal{C}_j$ comprising semantically similar images. To reduce manual burden, only $N_{per}$ images per cluster are manually annotated. Each cluster is scored based on the mini-batch of $N_{per}$ labeled images, yielding
\[
\bar{r}_j = \frac{1}{N_{per}} \sum_{i \in \mathcal{S}_j} r_i,  ~~\mathcal{S}_j \subset \mathcal{C}_j,
\]
where $r_i = 1$ denotes high relevance, $r_i = 0$ denotes low relevance, and $\mathcal{S}_j\subset\mathcal{C}_j$ is the $N_{per}$-image sample. Clusters with $\bar{r}_j \ge 0.8$ are assigned to the strong set $\mathcal{C}_{\text{s}}$, clusters with $0.2 < \bar{r}_j < 0.8$ form the mixed set $\mathcal{C}_{\text{m}}$, and clusters with $\bar{r}_j \le 0.2$ constitute the discard set $\mathcal{C}_{\text{d}}$.

Following this categorization, we can distill the salient features that should be retained, characterize mixed or ambiguous cases, and reliably identify instances that fail to meet the criteria and should be removed. 
By integrating human supervision with vision-language models, the original prompt
$\mathcal{P}_{\mathrm{init}}$ 
is optimized to obtain the improved prompt $\mathcal{P}_{\mathrm{opt}}$ .

\subsection{Mixture-of-Experts Knowledge Distillation}

\begin{figure}[t]
    \centering
    \includegraphics[width=\linewidth]{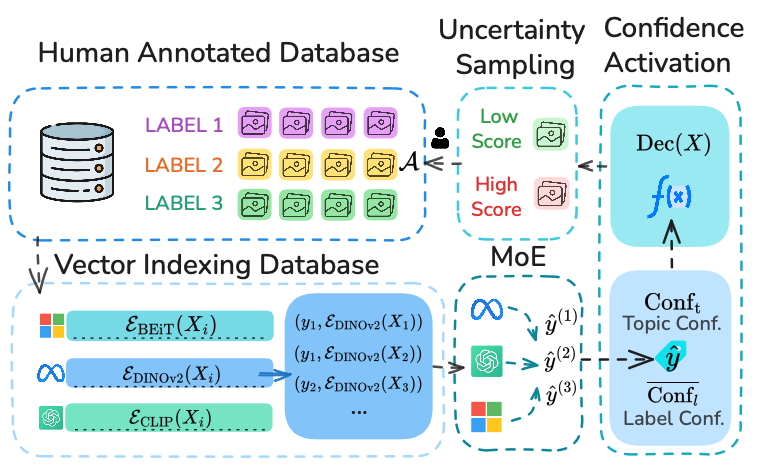}
    \caption{\textbf{Overall Structure of Mixture-of-Experts Knowledge Distillation.} Human-labeled images are embedded into a vector index. MoE predicts $\hat{y}$ with topic/label confidences; uncertainty sampling escalates low-confidence cases to annotators, and confidence activation decodes the final decision.}
    \label{fig:refine}
    \vspace{-15pt}
\end{figure}

\para{Overall Structure}
Mixture-of-Experts Knowledge Distillation (Figure~\ref{fig:refine}) refines the first-stage output $\Pi$ to extract images that meet dataset requirements. It has four parts: (1) constructing the vector-indexing database $\mathcal{I}$, (2) performing mixed-expert judgment, (3) applying Confidence Activation, and (4) filtering with selective manual labeling. Candidates are evaluated via $\mathcal{I}$, labeled by experts, validated by the activation module, and manually annotating a small routed subset to form a high-precision pool for the next round. After several iterations, images $X$ with predicted labels $\hat{y}$ in $\mathcal{Y}$ attain high accuracy; aggregating all processed data yields the dataset~$\pi$.

\vspace{0.5em}
\para{Vector Indexing Database}
The vector indexing database $\mathcal{I}$ is constructed from a small, manually annotated reference dataset $\mathcal{A} = \{(X_i, y_i)\}_{i=1}^N$, where $X_i$ denotes a labeled image and $y_i$ its corresponding category. To capture complementary information, three pre-trained image embedding models are employed: \textbf{CLIP}, \textbf{DINOv2}, and \textbf{BEiT}.

Specifically, CLIP provides semantic representations of the reference data 
$\mathcal{I}_{\mathrm{CLIP}} = \{(y_i, \mathcal{E}_{\mathrm{CLIP}}(X_i)) \mid X_i \in \mathcal{A}\}$, 
DINOv2 encodes spatial visual features 
$\mathcal{I}_{\mathrm{DINOv2}} = \{(y_i, \mathcal{E}_{\mathrm{DINOv2}}(X_i)) \mid X_i \in \mathcal{A}\}$, 
and BEiT captures universal visual representations 
$\mathcal{I}_{\mathrm{BEiT}} = \{(y_i, \mathcal{E}_{\mathrm{BEiT}}(X_i)) \mid X_i \in \mathcal{A}\}$. 
Each model embeds all annotated images, yielding three sub-databases that store category-vector pairs. 
The full vector-indexing database is then defined as the union of these sub-databases:
\[
\mathcal{I} = \{ \mathcal{I}_{\mathrm{CLIP}},  \mathcal{I}_{\mathrm{DINOv2}},  \mathcal{I}_{\mathrm{BEiT}} \}.
\]

\para{Mixture-of-Experts} The three experts $\mathcal{M}=\{\mathrm{CLIP},\mathrm{DINOv2},\mathrm{BEiT}\}$ query their respective sub-indices in the vector index to issue independent judgments. For each test image, expert $m$ embeds it and retrieves its top-$K$ neighbors, yielding
\[
\mathcal{N}^{(m)} = \{(\xi^{(m)}_k, y^{(m)}_k)\}_{k=1}^K, \quad m \in \mathcal{M},
\]
where $K\in\mathbb{N}$ is the number of retrieved neighbors, $\xi^{(m)}_k$ denotes the similarity between the test image and the $k$-th nearest neighbor in expert $m$’s sub-database, and $y^{(m)}_k$ is that neighbor’s category label. 

A temperature parameter $T$ converts similarities $\{\xi_k^{(m)}\}_{k=1}^K$ into
normalized softmax weights. The predicted label of expert $m$ is
\[
\hat{y}^{(m)} = \arg\max_{c \in \mathcal{Y}} 
\sum_{k=1}^K 
\frac{\exp(\xi_k^{(m)}/T)}{\sum_{j=1}^K \exp(\xi_j^{(m)}/T)}\,
\mathbf{1}[y_k^{(m)} = c],
\]
where $\mathbf{1}[\cdot]$ is the indicator function.

With three experts’ predictions $\hat{y}^{(m)}$, the final label $\hat{y}$ is obtained via majority voting. In the event of a three-way conflict, CLIP takes precedence and the image is also forwarded to the \textbf{Uncertainty Sampling} module, exploiting complementary semantic, structural, and general visual priors to improve the robustness and reliability of label assignment.

\vspace{0.5em}
\para{Confidence Activation}
For each test image, we assess the credibility of its assigned label $\hat{y}$ using two measures: topic confidence $\operatorname{Conf}_{\mathrm{t}}$ and label confidence $\operatorname{Conf}_{\mathrm{l}}$.

The topic confidence is the similarity of the Top-$K$ vectors from one expert,
\[
\mathrm{Conf}_{\mathrm{t}}(X) = \frac{1}{K}\sum_{k=1}^K \xi_k,
\]
capturing the affinity between $X$ and its retrieved neighbors.

The label confidence is the cosine similarity between the image embedding and the class-mean embedding,
\[
\mathrm{Conf}_{\mathrm{l}}^{(m)}(X)= \cos\big(\mathcal{E}^{(m)}(X), \boldsymbol{\mu}_{\hat{y}}^{(m)}\big),
\]
which quantifies the credibility of the predicted label $\hat{y}$.

The final label confidences are obtained by averaging the corresponding per-expert scores:
\[
\overline{\mathrm{Conf}_{\mathrm{l}}}(X)=\frac{1}{|\mathcal{M}|}\sum_{m\in\mathcal{M}}{\mathrm{Conf}_{\mathrm{l}}^{(m)}(X)}.
\]
Finally, the decision function $\mathrm{Dec}(X)$ is defined as:

\[
\resizebox{\columnwidth}{!}{$
\mathrm{Dec}(X)=
\begin{cases}
\hat{y}, & \text{if } \mathrm{Conf}_{\mathrm{t}}(X)\ge\theta_{\mathrm{t}} \ \text{and } 
                 \overline{\mathrm{Conf}_{\mathrm{l}}}(X)\ge\theta_{\mathrm{l}},\\
\text{non-target}, & \text{otherwise}.
\end{cases}
$}
\]

\para{Uncertainty Sampling}
To reduce labeling cost and maximize supervised-data utility, we adopt an \textit{Uncertainty Sampling} module that prioritizes informative samples. For each predicted category, we compute a Feature Alignment Score (FAS)—the cosine similarity between the image embedding and the class-mean embedding $\boldsymbol{\mu}_{\hat{y}}$ in the index—averaged over experts:

\[
\overline{\mathrm{FAS}}(X,\hat{y}) = \frac{1}{|\mathcal{M}|}\sum_{m\in\mathcal{M}} \frac{\mathcal{E}^{(m)}(X)\cdot \boldsymbol{\mu}_{\hat{y}}^{(m)}}{\|\mathcal{E}^{(m)}(X)\|\,\|\boldsymbol{\mu}_{\hat{y}}^{(m)}\|}.
\]
Samples are ranked per category by this score.
Based on the ranking, we form a low-score pool (bottom $\alpha\%$ within each category ${c_k}$), from which $K_L$ samples are randomly drawn, reflecting higher uncertainty.

 For each non-target sample $X$, we compute its alignment scores with respect to all classes:

\[
\overline{\mathrm{FAS}}(X, c_k),
\quad c_k \in \{c_1, c_2, \dots, c_{|\mathcal{Y}|}\}.
\]

where $|\mathcal{Y}|$ denotes the total number of classes. We then define the boundary strength and boundary attribution category of $X$ as
\[
B(X) = \max_{k=1,\dots,|\mathcal{Y}|}\,\overline{\mathrm{FAS}}(X,c_k),
\]
\[
y^\star(X) = \arg\max_{k=1,\dots,|\mathcal{Y}|}\,\overline{\mathrm{FAS}}(X,c_k).
\]
$B(X)$ measures the strongest class-wise alignment and $y^\star(X)$ identifies the aligned class; non-targets are ranked by $B(X)$ and the top $K_H$ are taken as boundary candidates. Samples from both pools are then manually annotated, and the resulting labels are integrated into the Vector Indexing Database to enable the next expert iteration.

\subsection{Region-Evidence VLM Adversarial Labeling}

\begin{figure}[h]
    \centering
    \includegraphics[width=0.9\linewidth]{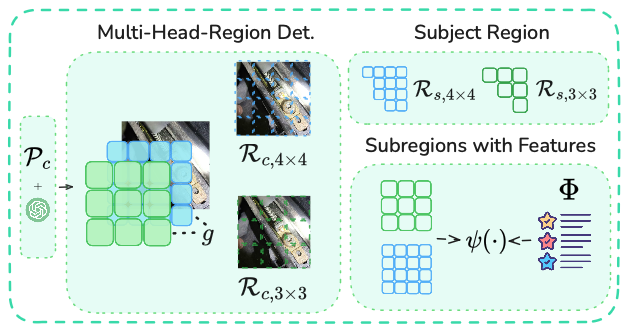}
    \caption{\textbf{Architecture of the VLM Proposer.} The Proposer outputs Multi-Head-Region grids, isolates the subject and subregions, and annotates each with semantic cues $\boldsymbol{\Phi}$.}
    \label{fig:proposer}
\end{figure}

\begin{figure}[h]
    \centering
    \includegraphics[width=0.9\linewidth]{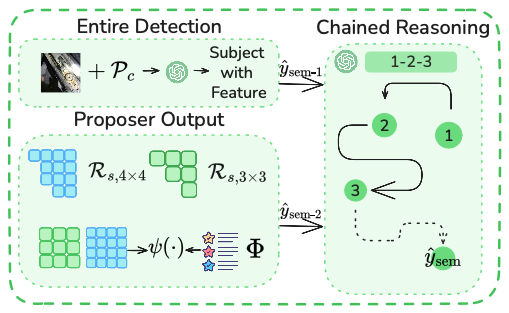}
    \caption{\textbf{Architecture of the VLM Validator.} The validator performs global inference, fuses evidence via chained reasoning, and outputs $\hat{y}_{\text{sem}}$.}
    \label{fig:validator}
    \vspace{-15pt}
\end{figure}

\para{Overall Workflow} Building on large-scale acquisition/filtering and mixture-of-experts labeling, the process yields strong but imperfect labels for dataset~$\pi$. To further refine accuracy and semantic granularity, we introduce a two-stage, Region-Evidence VLM (RE-VLM): a \textit{VLM Proposer} (Figure~\ref{fig:proposer}) extracts global and multi-region features and a \textit{VLM Validator} (Figure~\ref{fig:validator}) infers and verifies categories and attributes from these cues, yielding image-specific, proprietary labels. 

\vspace{-0.5em}
\para{VLM Proposer}  
As the information proposer, the Proposer first resolves the image subject. In edge settings with limited priors, category labels are insufficient. Therefore, conditioned on the label, we compose general yet discriminative class-level feature descriptions to build the VLM prompt $\mathcal{P}_c$. For labels with richer semantics, we sharpen spatial focus via Multi-Head-Region detection over the subject, providing finer evidence for the Validator. Concretely, we use two heads ($3\times3$ and $4\times4$). The set of subregions at granularity $g$ for class $c$ is defined as
\[
\mathcal{R}_{c,g} = \{ \mathbf{r}_{c,g}^{(1)}, \mathbf{r}_{c,g}^{(2)}, \dots, \mathbf{r}_{c,g}^{(N_g)} \}, \quad g \in \{\text{3$\times$3}, \text{4$\times$4}\}
\]
where $\mathcal{R}_{c,g}$ denotes the collection of subregions obtained at granularity $g$, and $N_g$ is the corresponding number of subregions.

Finally, the model is provided with the class feature description, the input image $X$, the subject region (denoted below as $\mathcal{R}_{s,g}$), and all subregions whose features lie in the feature set $\boldsymbol{\Phi}$:

\[
\mathsf{Prop}(X,\mathcal{P}_c)\to\big(\mathcal{R}_{s,g},\{\psi(\mathbf{r}_{c,g}^{(j)},\boldsymbol{\Phi})\}_{j=1}^{N_g}\big),\ \mathcal{R}_{s,g}\subset\mathcal{R}_{c,g},
\]
where $\psi(\cdot,\boldsymbol{\Phi})$ indicates whether a subregion $\mathbf{r}_{c,g}^{(j)}$ contains features from $\boldsymbol{\Phi}$.

\vspace{0.5em}
\para{VLM Validator} The Validator performs whole-image inference to produce $\hat{y}_{\text{sem-1}}$ under $\mathcal{P}_c$, then ingests the main region and multi-granularity subregions from the Proposer, aligning them with $\mathcal{P}_c$ to obtain $\hat{y}_{\text{sem-2}}$.
For reliability, we use a chained consistency check: detecting features in $\boldsymbol{\Phi}$ and inferring their presence; using Proposer outputs to localize/evaluate sub-granularity regions against $\mathcal{R}_{c,g}$; and fusing both pieces of evidence to yield $\hat{y}_{\text{sem}}$.

\section{Experiment}

\begin{table*}[ht]
\centering
\vspace{-7pt}
\caption{\textbf{Method Comparison for Flooded-Car Study.} Contrasts methods across both stages to assess overall pipeline improvements. ``{-}'' indicates unavailable metrics in that stage (not applicable, failed, or not reported). Retrieval-Augmented GPT-5 uses the same human-annotation budget as ReCCur.}
\label{tab:method_comparison_two_block}
\begingroup
\footnotesize
\setlength{\tabcolsep}{6pt}            
\renewcommand{\arraystretch}{1.15}     
\newcommand{\NA}{\textemdash}          
\begin{tabular*}{\textwidth}{@{\extracolsep{\fill}} lcccccccccc @{}}
\toprule
\textbf{Method} &
\multicolumn{5}{c}{\textbf{Image Recognition and Noise Filtering Stage}} &
\multicolumn{4}{c}{\textbf{Semantic Labeling Stage}} \\
\cmidrule(lr){2-6}\cmidrule(lr){7-10}
& \textbf{Precision} & \textbf{Recall} & \textbf{F1 score} & \textbf{NRR} & \textbf{CDRR}
& \textbf{Precision} & \textbf{Recall} & \textbf{F1 score} & \textbf{Perfect match} \\
\midrule
KNN                            & 0.788 & 0.898 & 0.832 & 0.785 & 0.982 & \NA    & \NA    & \NA    & \NA \\
K\mbox{-}means                 & 0.811 & 0.683 & 0.691 & 0.497 & 1.000 & \NA    & \NA    & \NA    & \NA \\
Vo et al.~\cite{article17}
                               & 0.886 & 0.576 & 0.685 & 0.777 & 0.815 & \NA    & \NA    & \NA    & \NA \\
YOLO\mbox{-}World v2\cite{Cheng2024YOLOWorld}
                               &  0.253 & 0.129 & 0.094 & 0.968 & 0.032   & \NA    & \NA    & \NA    & \NA \\
SigLIP2\cite{Tschannen2025SigLIP2}                         & 0.748 & 0.474 & 0.393 & 0.412 & 1.000 & \NA    & \NA    & \NA    & \NA \\
CaSED\cite{Conti2023CaSED}                          & 0.774 & 0.583 & 0.481 & 1.000 & 0.173 & \NA    & \NA    & \NA    & \NA \\
Qwen\mbox{-}VL\,2.5 (Instruction QA)
                               & 0.846 & 0.799 & 0.783 & 0.784 & 0.831 & 0.520  & 0.830  & 0.506  & 0.182 \\
GPT\mbox{-}5 (Instruction QA)  & 0.892 & 0.863 & 0.846 & 0.839 & 0.886 & 0.718    &  0.884    & 0.776    & 0.649 \\
Retrieval\mbox{-}Augmented GPT\mbox{-}5
                               & 0.902 & 0.885 & 0.873 & 0.852 & 0.879 & 0.725 & 0.870 & 0.801    & 0.661 \\
\textbf{ReCCur (ours)}                  & \textbf{0.954} & \textbf{0.972}  & \textbf{0.963} & \textbf{0.956} & \textbf{0.990} & \textbf{0.883}  & \textbf{0.851}  & \textbf{0.867}  & \textbf{0.801} \\
\bottomrule
\end{tabular*}
\endgroup
\vspace{-10pt}
\end{table*}

In this section, we verify the effectiveness of our entire architecture through detailed experiments. Experiments are designed to test the feasibility of the architecture under different conditions by constructing a new dataset of flooded-car corner cases, psychedelic mushroom corner cases and Wall-Damage corner cases.

\subsection{Implementation}
\paragraph{Setup} In the experiments, all VLM-related components used ChatGPT-5 (public API; checkpoint: 2025-08-07). We employed three pre-trained visual embedding models—CLIP ViT-L/14, DINOv2-L, and BEiT-L/16—as the visual backbones across all comparative studies. For each backbone, we built a separate retrieval database to produce expert labels. To ensure feasibility under modest computational resources, the end-to-end pipeline was run primarily on a laptop with an NVIDIA GeForce RTX 3060 (6 GB) paired with an Intel Core i7-11800H CPU; workloads impractical on this machine were executed on a server-class NVIDIA A10 GPU.

\vspace{-1em}
\paragraph{Evaluation metrics.} For the Mixture-of-Experts Knowledge Distillation, we evaluate all methods with macro precision, recall, and F1—computed from TP/FP/FN and, unless stated otherwise, macro-averaged across classes—and with two curation-specific measures: Noise Removal Rate (NRR), the fraction of human-verified noisy samples eliminated, and Clean Data Retention Rate (CDRR), the fraction of human-verified clean samples preserved. For Region-Evidence VLM Adversarial Labeling, we additionally evaluate the predicted labels using precision and recall to validate semantic feature fidelity.
\vspace{-1em}
\paragraph{Baseline Selection.} We compare our approach with four representative families to cover metric learning, noise pruning, open-vocabulary recognition, and language-guided reasoning. Implementations are taken from official repositories and public releases to ensure fairness and reproducibility. All baselines are run under identical configurations—data splits, input resolution, prompts/text vocabularies, and compute budget—following our evaluation protocol and metrics (Precision/Recall/F1, NRR, CDRR). This design enables comparison across cluster-based methods, open-vocabulary models, and VLM QA variants.

\subsection{Case Study: Flooded-Car Inspection}
\begin{figure}[h]
    \centering
    \includegraphics[width=\linewidth]{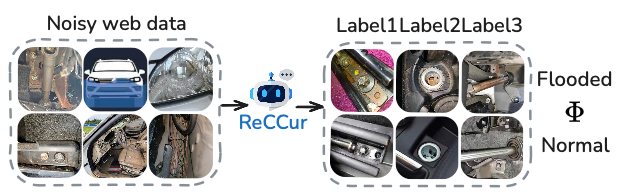}
    \caption{\textbf{Task Configuration and Objectives.} ReCCur generates fine-grained part-level semantic labels and enables comparison of different methods in recognition and filtering performance.}
    \label{fig:task}
\end{figure}

\para{Background}
Extreme weather has made vehicle flooding increasingly common, and in some markets flood-damaged cars are refurbished and resold. These vehicles compromise safety and consumer protection, while buyers often lack the expertise to recognize water-damage indicators. Therefore, we employ ReCCur to construct a flooded-vehicle inspection-point dataset aligned with diagnostic parts recognized by accredited institutions. ReCCur distills high-quality samples from noisy web data and generates part-specific, feature-aware semantic labels, yielding a dedicated dataset and enabling from-scratch validation in edge scenarios. Figure~\ref{fig:task} outlines the task configuration and experimental objectives. 

\vspace{0.5em}
\para{Implementation Details}
Drawing on flooded-vehicle inspection reports, we select common trace points as seeds to build $\mathcal{K}_{\mathrm{V}}$ and $\mathcal{K}_{\mathrm{T}}$, forming $\mathcal{K}$ for raw data acquisition; multilingual, multi-channel web crawling yields $\mathcal{D}$. A multimodal similarity threshold $\tau_{\mathrm{mm}}=0.4$ (e.g., weights: 0.5 img–desc, 0.3 desc–kw, 0.2 img–kw) is applied to filter noisy samples before clustering. Using Multimodal Data Filtering and the Human Supervision Loop, we manually resolve $N_{per}{=}5$ images per cluster, obtaining $\Pi$ under prompt $\mathcal{P}_{\mathrm{opt}}$. The label space $\mathcal{Y}$ includes seven immersion points (A: seat belt, B: seat track, C: wiring harness, D: spare tire well, E: foam padding, F: 12V power outlet, G: steering column) and a noise class H. In Mixture-of-Experts Distillation, we set $K\!=\!7$, $\alpha\!=\!20\%$, $K_L\!=\!3$, and $K_H\!=\!3$, with thresholds $\theta_{\mathrm{t}}{=}0.65$ and $\theta_{\mathrm{l}}{=}0.45$, to derive $\pi$. For VLM feature detection, $\boldsymbol{\Phi}$ contains “rust,” “dust and sand,” “mold,” “aged,” and “none.” All categories and traces are manually annotated once, after which the pipeline generates semantic labels. More detailed task descriptions are provided in the supplementary material.

The first stage comprised twelve iterations, and the second stage five. Validation annotations were conducted by annotators with automotive expertise, following a unified evaluation standard to ensure consistent labels and inter-annotator agreement.

\begin{table*}[t]
\centering
\caption{\textbf{Ablation Study of ReCCur Components.} Isolates the impact of multi-experts, Confidence Activation, and Region-Evidence.}
\vspace{-5pt}
\label{tab:ablation}
\footnotesize
\setlength{\tabcolsep}{6pt}
\renewcommand{\arraystretch}{1.15}
\newcommand{\NA}{\textemdash}
\begin{tabular*}{\textwidth}{@{\extracolsep{\fill}} cccc ccccc cccc @{}}
\toprule
\multicolumn{4}{c}{\textbf{Modules}} &
\multicolumn{5}{c}{\textbf{Image Recognition and Noise Filtering Stage}} &
\multicolumn{4}{c}{\textbf{Semantic Labeling Stage}} \\
\cmidrule(lr){1-4}\cmidrule(lr){5-9}\cmidrule(lr){10-13}
\textbf{MoE} & \textbf{CA} & \textbf{VLM} & \textbf{RE-VLM} &
\textbf{Precision} & \textbf{Recall} & \textbf{F1} & \textbf{NRR} & \textbf{CDRR} &
\textbf{Precision} & \textbf{Recall} & \textbf{F1} & \textbf{Perfect Match} \\
\midrule
\xmark & \xmark & \xmark & \xmark &
0.807 & 0.864 & 0.832 & 0.904   & 0.906   & \NA   & \NA   & \NA   & \NA \\
\cmark & \xmark & \xmark & \xmark &
0.880 & 0.872 & 0.876 & 0.912 & 0.925 & \NA   & \NA   & \NA   & \NA \\
\cmark & \cmark & \xmark & \xmark &
0.954 & 0.972  & 0.963 & 0.956 & 0.990   & \NA   & \NA   & \NA   & \NA \\
\cmark & \cmark & \cmark & \xmark &
0.954 & 0.972  & 0.963 & 0.956 & 0.990 & 0.731  & 0.812 & 0.769 & 0.659 \\
\cmark & \cmark & \xmark & \cmark 
& 0.954 & 0.972  & 0.963 & 0.956 & 0.990 & 0.894  & 0.859 & 0.856  & 0.802 \\

\bottomrule
\end{tabular*}
\vspace{-0.6em}
\end{table*}

\begin{table*}[t]
\centering
\caption{\textbf{Method Comparison for the Psychedelic Mushroom Study.} GPT-based methods vs. ReCCur (compact).}
\vspace{-5pt}
\label{tab:secondcase}
\footnotesize
\setlength{\tabcolsep}{3pt}
\newcommand{\NA}{\textemdash}
\begin{tabular*}{\textwidth}{@{\extracolsep{\fill}} lcccccc @{}}
\toprule
\textbf{Method} &
\multicolumn{3}{c}{\textbf{Image Recognition \& Noise Filtering}} &
\multicolumn{3}{c}{\textbf{Semantic Labeling}} \\
\cmidrule(lr){2-4}\cmidrule(lr){5-7}
& \textbf{P/R/F1} & \textbf{NRR} & \textbf{CDRR}
& \textbf{Color P/R/F1} & \textbf{Shape P/R/F1} & \textbf{Perfect} \\
\midrule
GPT\mbox{-}5                        & 0.312/0.257/0.266 & 0.236 & 0.354 & 0.726/0.748/0.735 & 0.753/0.701/0.714 & 0.782 \\
Retrieval\mbox{-}Aug.\ GPT\mbox{-}5 & 0.569/0.337/0.356 & 0.138 & 0.410 & 0.738/0.735/0.736 & 0.747/0.692/0.702 & 0.774 \\
\textbf{ReCCur (ours)}                       & \textbf{0.947/0.941/0.941} & \textbf{0.984} & \textbf{0.916} & \textbf{0.940/0.892/0.913} & \textbf{0.816/0.824/0.800} & \textbf{0.875} \\
\bottomrule
\end{tabular*}
\vspace{-7.5pt}
\end{table*}

\begin{figure}[h]
    \centering
    \begin{minipage}[t]{0.495\linewidth}
        \centering
        \includegraphics[width=\linewidth]{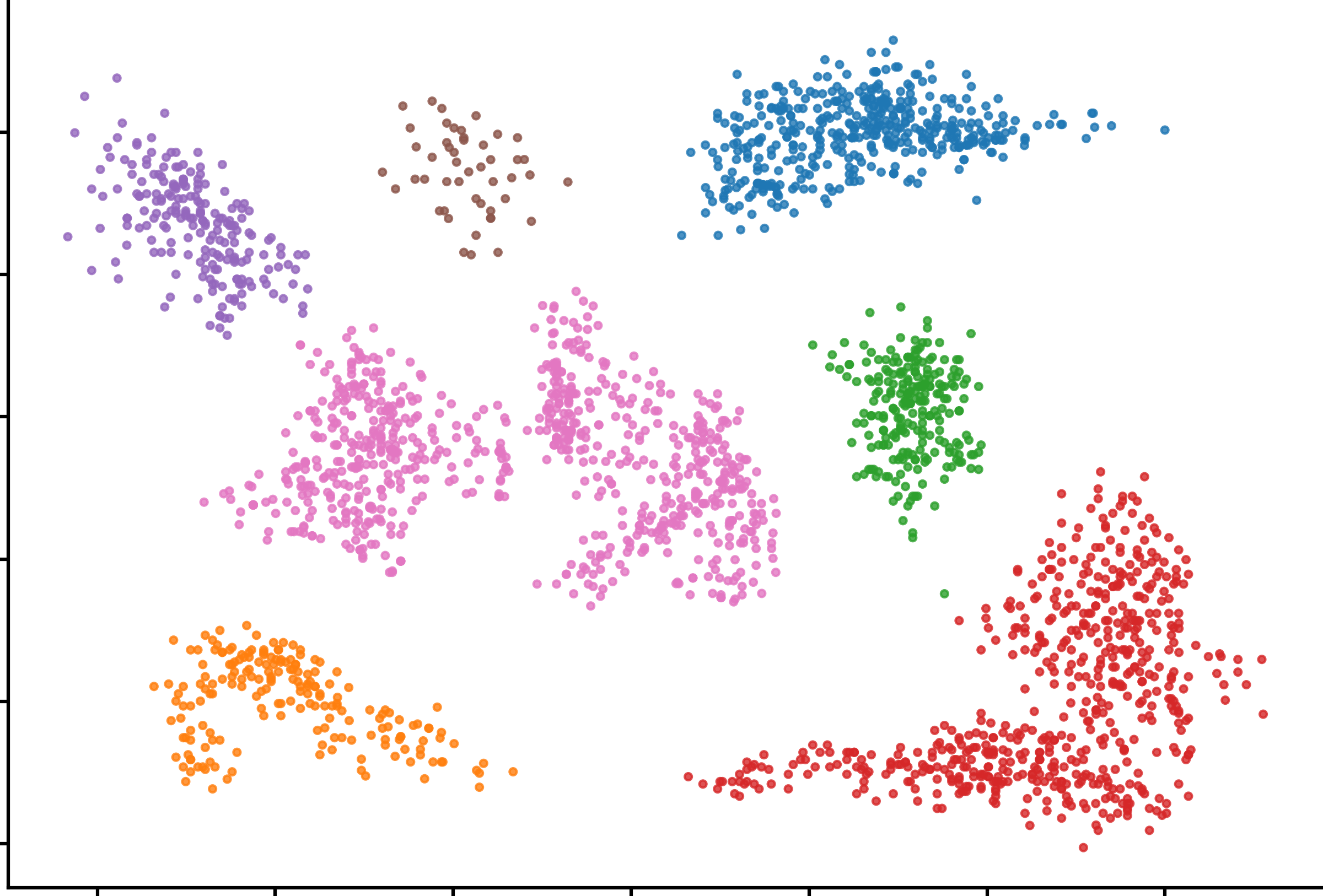}
        \vspace{0.25em}
        \small (a) High similarity (\(\ge \tau_{\mathrm{mm}}\)).
    \end{minipage}\hfill
    \begin{minipage}[t]{0.495\linewidth}
        \centering
        \includegraphics[width=\linewidth]{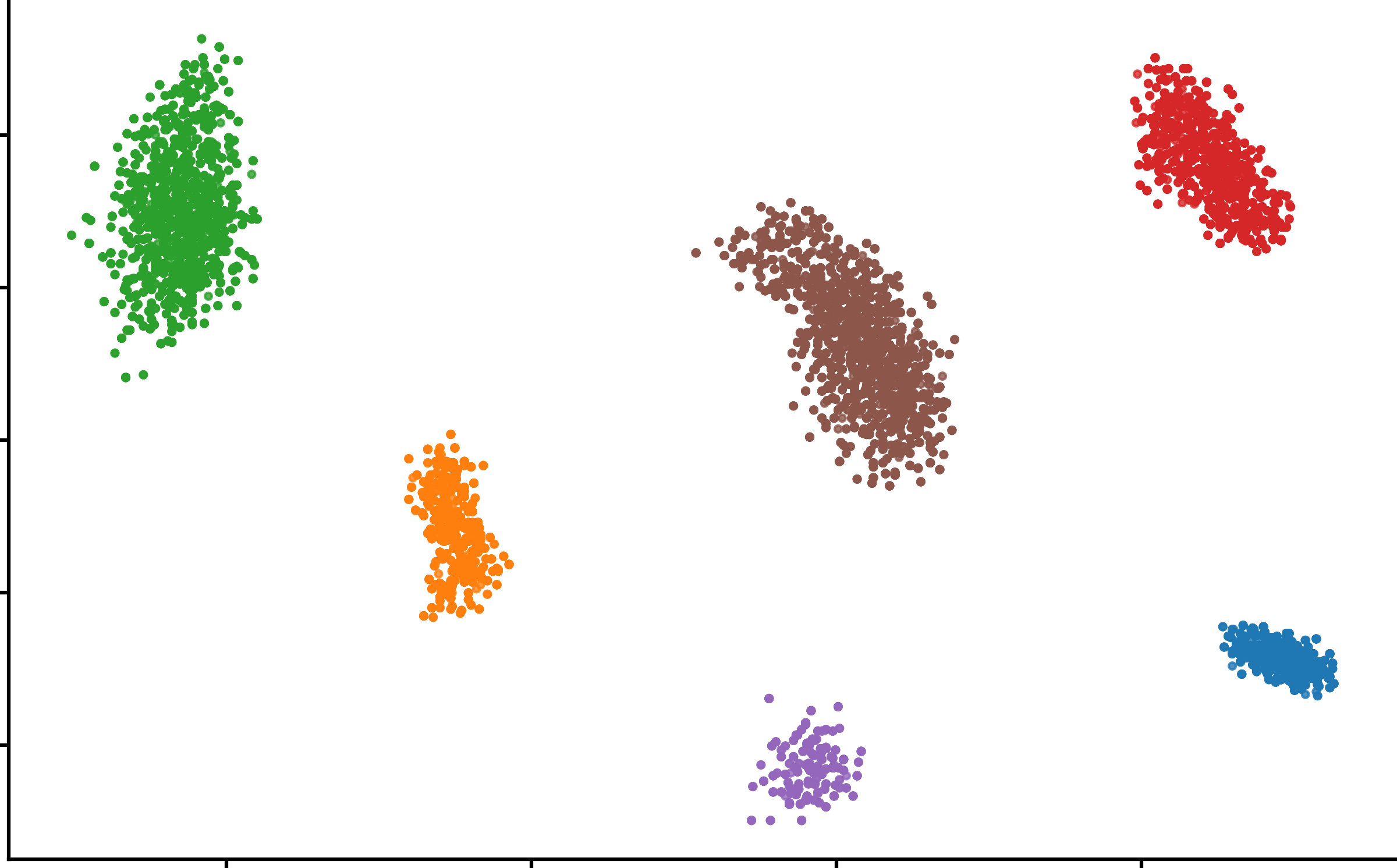}
        \vspace{0.25em}
        \small (b) Low similarity (\(< \tau_{\mathrm{mm}}\)).
    \end{minipage}
    \caption{\textbf{Clustering Results for Data Filtering}
    Split by multi-modal similarity threshold \(\tau_{\mathrm{mm}}\) before clustering (HDBSCAN).}
    \label{fig:cluster-singlecol}
    \vspace{-0.6em}
\end{figure}

\begin{figure}[ht]
    \centering
     \includegraphics[width=\linewidth]{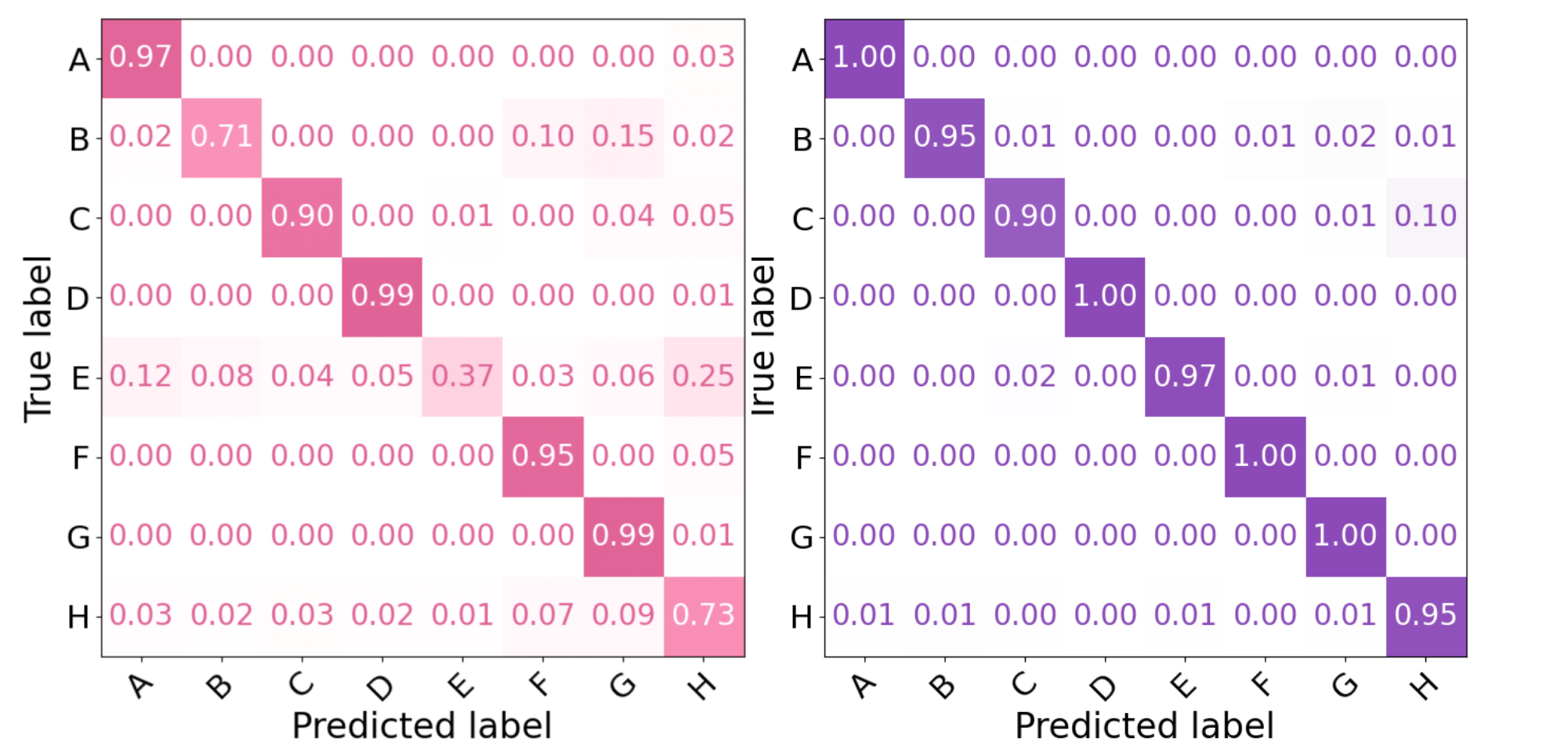}
    \caption{\textbf{Confusion matrices for Flooded-Car Study} the first round (left) and final round (right).}
    \label{fig:confusion matrix1}
    \vspace{-10pt}
\end{figure}

\vspace{-0.5em}
\para{Result}
Across rounds, the first stage yields clear inter-cluster separation and compact intra-cluster structure (Figure~\ref{fig:cluster-singlecol}), indicating effective pruning of off-topic and low-evidence samples. Quantitatively, ReCCur attains the highest identification accuracy while preserving data utility: both NRR and CDRR exceed 0.9 (Table~\ref{tab:method_comparison_two_block}), with only ~4\% of samples escalated for manual review over two stages. The confusion matrices exhibit strong diagonal dominance and uniformly low off-diagonal mass (Figure~\ref{fig:confusion matrix1}), evidencing low cross-class confusion and stable behavior under limited supervision.

\subsection{Ablation Experiments}
We conducted ablation experiments at each stage of the architecture to verify their effectiveness.

\begin{figure}[h]
    \centering
    \includegraphics[width=\linewidth]{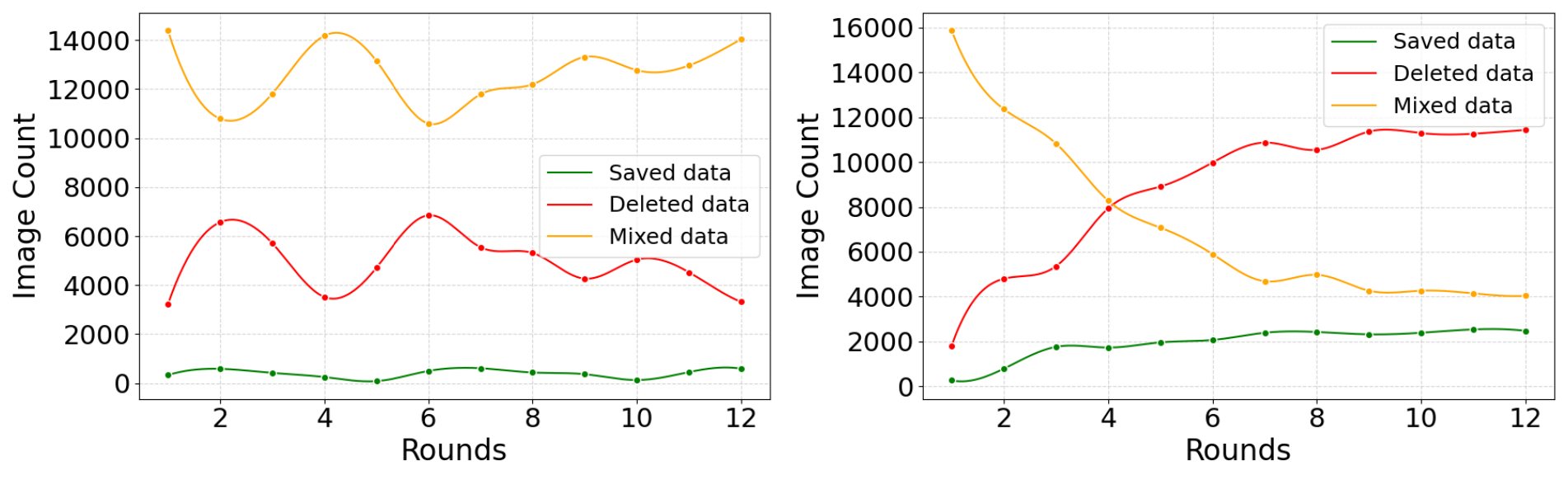}
    \vspace{-15pt}
    \caption{\textbf{Comparison of Image Counts Across Sets After (left) and Before (right) Ablation}}
    \label{fig:linechart2}
    \vspace{-15pt}
\end{figure}

\vspace{0.5em}
\para{Large-Scale Data Acquisition and Filtering}
In this stage, we ablated the multimodal similarity filtering, removing multimodal similarity and augmented spatial representations $\mathbf{z}_i$ and using only image embeddings $\mathcal{E}_{\mathrm{CLIP}}$ to verify performance. Figure~\ref{fig:linechart2} demonstrates that the three sets fluctuate greatly after ablation, making convergence difficult.

\vspace{0.5em}
\para{Mixture-of-Experts Knowledge Distillation} In this stage, we ablate (i) single-expert ($|\mathcal{M}|=1$) vs.\ multi-expert ($|\mathcal{M}|=3$), (ii) Confidence Activation on/off. As shown in Table~\ref{tab:ablation}, the MoE and CA configuration improves detection and strengthens both noise removal and clean-data retention. 

\vspace{0.5em}
\para{Region-Evidence VLM Adversarial Labeling}
We validate this stage by comparing a standalone VLM with a Region-Evidence-augmented VLM. The results indicate that incorporating Region-Evidence enhances trace localization, mitigates confusion between subject and background, and yields more accurate semantic labels. More detailed task descriptions are provided in the supplementary material.

\subsection{Case Study: Psychedelic Mushroom}
\para{Background} In some regions, foraging and consuming wild mushrooms remain common. Poisoning cases frequently involve hallucinations—often described as seeing “rainbow figures”—and the large number of such incidents has drawn significant online attention. This further underscores the need for reliable toxic-mushroom identification.

\vspace{0.5em}
\para{Implementation} To better stress noisy web cases, we further build a small “psychedelic-mushroom” subset: many web pages about mushrooms mix real toxic species with meme-like or highly stylized images (bright, rainbow-colored, “LSD-like”). Visually, these samples are far from real mushrooms but they are still retrieved by text queries such as “toxic mushroom” or “magic mushroom”, making them typical corner cases for multimodal curation. In our experiments we group mushroom images into four coarse categories: (i) realistic deadly/toxic mushrooms, (ii) normal/edible mushrooms, (iii) conditionally edible mushrooms and (iv) psychedelic or meme-like mushroom (noise) images from the web. We also use Qwen\mbox{-}VL\,2.5 as the RE\mbox{-}VLM backend to verify stage\mbox{-}3 generality. 

\vspace{0.5em}
\para{Result}
With only two rounds and 67 human labels, ReCCur surpasses all other methods across all metrics (Table~\ref{tab:secondcase}).
Replacing the VLM with Qwen\mbox{-}VL\,2.5 in the RE\mbox{-}VLM stage yields consistent gains in semantic recall and perfect\mbox{-}match accuracy, confirming model\mbox{-}agnostic generality under low\mbox{-}supervision. ReCCur transfers to the mushroom domain without redesign, though performance hinges on trace-level cues in the VLM embeddings.

\vspace{0.5em}
\subsection{Case Study: Wall-Damage Detection}
We further test ReCCur on wall-damage detection; it remains accurate, separates wall types, and detects fine traces. See the Supplementary Material for details and full results.

\section{Conclusion}
We introduced \textbf{ReCCur}, a modular curation–labeling framework integrating multimodal filtering, mixture-of-experts distillation, and region-evidence VLM verification. On flooded-vehicle and mushroom studies, ReCCur achieves high recognition accuracy, strong noise suppression, and high clean-data retention with minimal human effort. Standardized prompts/embeddings with confidence activation yield stable, auditable decisions and efficient uncertainty routing. The resulting datasets exhibit clear structure and support higher downstream performance, demonstrating the framework’s cross-domain applicability and practical deployability.


\clearpage
\setcounter{page}{1}
\maketitlesupplementary


\section{List of Symbols}
To make the subsequent derivations easier to follow, we summarize the notations used throughout this section in Table~\ref{tab:SymbolsASD}.

\section{VLM Baselines Configuration}
The detailed parameters of the VLM we used are provided in Table~\ref{tab:baselines_config}.
\begin{table*}[t]
\centering
\caption{\textbf{VLM Baselines Configuration.} Summary of model, supervision style, temperature setting, and retrieval usage for all VLM-based baselines under the same human supervision budget.}
\label{tab:baselines_config}
\footnotesize
\resizebox{\textwidth}{!}{%
\begin{tabular}{lcccccc}
\toprule
\textbf{Method} & \textbf{Model} & \textbf{Details} & \textbf{Train} & \textbf{Temp} & \textbf{Retr.} \\
\midrule
GPT-based QA & GPT-5 & Q/A (single-stage) & No & Default Val. & No \\
Retrieval-Aug. GPT-5 & GPT-5 + CLIP index & Retrieval Q/A & No &  Default Val. & Yes  \\
\textbf{ReCCur (ours)} & GPT-5/Qwen\mbox{-}VL\,2.5 + CLIP/DINOv2/BEiT & MoE + Region evidence & No &  Default Val. & Yes  \\
\bottomrule
\end{tabular}%
}
\end{table*}

\section{Hardware Processing Speed}
Table~\ref{tab:full_perf_table} presents the inference performance of the ReCCur framework across various hardware platforms using a train-free PyTorch implementation. On an Intel i7-11700 CPU, ReCCur achieves 1.6 images per second, serving as the baseline. The NVIDIA RTX 3060 improves throughput to 4.1 images per second with 5.6GB of memory usage, demonstrating solid performance on a consumer-grade GPU. On the data center-grade NVIDIA A10 GPU, ReCCur processes 8.2 images per second with a batch size of 16 and 5.9GB memory usage. The limited throughput gain on A10 suggests that ReCCur is not heavily compute-bound and exhibits minimal GPU parallelism overhead.
\begin{table*}[t]
\begin{center}
\caption{\textbf{Inference Performance Comparison on CPU and GPU Platforms.} Results are obtained using a PyTorch-based train-free model.}
\label{tab:full_perf_table}
\resizebox{\textwidth}{!}{%
\begin{tabular}{@{}lcccccccc@{}}
\toprule
\textbf{Device} & \textbf{Framework} & \textbf{Batch Size} & \textbf{Speed (img/s)}  & \textbf{GPU Mem (GB)} &  \textbf{Notes} \\
\midrule
Intel i7-11700 CPU & PyTorch & 1 & 1.6  & --  & Baseline CPU performance \\
Intel i7-11700 CPU+NVIDIA RTX 3060 GPU & PyTorch & 16& 4.1 & 5.6  & Consumer-grade GPU \\
Intel Xeon(Ice Lake) Platinum 8369B+NVIDIA A10 GPU & PyTorch & 16 & 8.2 & 5.9 & Data center card \\

\bottomrule
\end{tabular}%
}
\end{center}
\vspace{-1em}
\end{table*}
\section{Flooded-Car Study Details}
\subsection{Task descriptions}
\begin{figure*}[t]
    \centering
    \includegraphics[width=\textwidth]{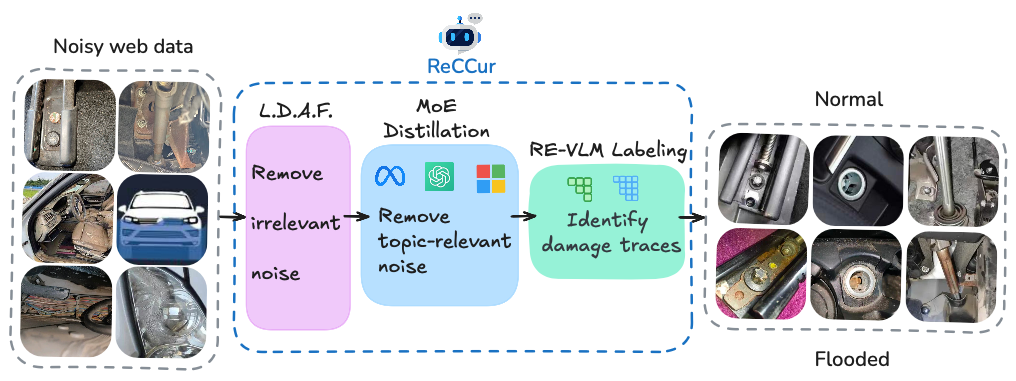}
    \caption{\textbf{Details of task and pipeline.} Starting from noisy web-crawled images, the Large-Scale Data Acquisition and Filtering (L.D.A.F.) stage removes visually irrelevant and off-topic samples. The Mixture-of-Experts (MoE) Distillation stage further eliminates topic-relevant noise and assigns reliable coarse labels. Finally, the Region-Evidence VLM (RE-VLM) module performs fine-grained reasoning to identify damage traces—such as corrosion, mud residues, rusted bolts, and wiring-harness oxidation—allowing the system to distinguish normal vehicles from flood-damaged ones.}
    \label{fig:task detail}
\end{figure*}

Flood-damaged vehicles exhibit subtle visual traces on metal parts and hidden compartments. In our dataset, each region is labeled as one of five trace types: “rust”, “dust and sand”, “mold”, “aged”, and “none”. We define “aged” as heavy damage where visible traces cover more than 50\% of the main structure. ReCCur takes noisy web images as input and, through the three stages in Figure~\ref{fig:task detail}, progressively removes off-topic and unreliable samples while assigning these fine-grained trace labels and the final normal/flooded status.

\subsection{Detailed Results}
\begin{figure}[t]
    \centering
    \includegraphics[width=0.9\linewidth]{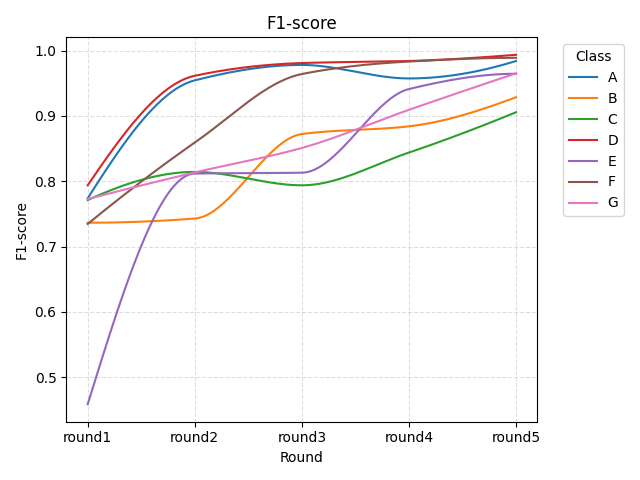}
    \caption{\textbf{F1-Score Progression over Iterative Rounds.} Per-class F1-scores (A--G) rise steadily across five rounds.}
    \label{fig:linef1}
\end{figure}

As shown in Figure~\ref{fig:linef1} and Table~\ref{tab:cls_metrics_no_support}, ReCCur achieves strong per-class results: all F1-scores \(\ge 0.9\). Best classes are D (F1 = 0.9936), F (0.9890), and A (0.9841). Four classes reach perfect recall (A/D/F/G = 1.0000); among them, G has lower precision (0.9327), indicating more false positives. The weakest class is C (F1 = 0.9057) due to slightly lower recall (0.8955). Overall precision is high across classes.

\begin{table}[h]
\centering
\caption{\textbf{Per-class precision/recall/F1 of Flooded-Car Study.}}
\label{tab:cls_metrics_no_support}
\small
\setlength{\tabcolsep}{6pt}
\renewcommand{\arraystretch}{1.15}
\begin{tabular}{lccc}
\toprule
\textbf{Class} & \textbf{Precision} & \textbf{Recall} & \textbf{F1-score} \\
\midrule
A & 0.9686 & 1.0000 & 0.9841 \\
B & 0.9070 & 0.9512 & 0.9286 \\
C & 0.9160 & 0.8955 & 0.9057 \\
D & 0.9873 & 1.0000 & 0.9936 \\
E & 0.9550 & 0.9745 & 0.9646 \\
F & 0.9781 & 1.0000 & 0.9890 \\
G & 0.9327 & 1.0000 & 0.9652 \\
H & 0.9891 & 0.9569 & 0.9727 \\
\bottomrule
\end{tabular}
\vspace{-0.5em}
\end{table}

\begin{table}[h]
\centering
\caption{\textbf{Trace-type results of Flooded-Car Study.}}
\label{tab:trace_type_metrics1}
\small
\setlength{\tabcolsep}{8pt}
\renewcommand{\arraystretch}{1.15}
\begin{tabular}{lccc}
\toprule
\textbf{Trace Type} & \textbf{Precision} & \textbf{Recall} & \textbf{F1-Score} \\
\midrule
rust       & 0.9892 & 0.9387 & 0.9633 \\
dust/sand  & 0.9845 & 0.5134 & 0.6749 \\
aged       & 0.7263 & 0.9452 & 0.8214 \\
mold       & 0.9836 & 0.9231 & 0.9524 \\
none       & 0.7871 & 0.9720 & 0.8698 \\
\midrule
Macro Avg & 0.8942 & 0.8585 & 0.8564 \\
\bottomrule
\end{tabular}
\end{table}

Overall performance is stable (Micro F1 = 0.8669, Macro F1 = 0.8564) but class imbalance is evident. Dust/sand achieves high precision (0.9845) yet low recall (0.5134), indicating many missed positives. Aged shows high recall (0.9452) but lower precision (0.7263), suggesting over-prediction and confusion with none or rust. None has very high recall (0.9720) but moderate precision (0.7871), implying overuse of the “no-trace” label (Table~\ref{tab:trace_type_metrics1}). Rust and mold remain balanced and strong. These results suggest improving fine-texture cues for dust/sand and adjusting thresholds or regional evidence for aged vs. none could enhance class balance. Notwithstanding the generally satisfactory results, reduced resolution may compress feature expressiveness and slightly impact performance; sensitivity to sub-pixel traces remains improvable.

As summarized in Table~\ref{tab:crawl_stats}, our initial web crawl yields 17,943 images, of which only about 10\% are true target samples, while roughly 60\% are loosely topic-related and 30\% are clearly off-topic. This skewed composition highlights the necessity of a dedicated curation pipeline to suppress noise and recover a clean, high-precision dataset before downstream training.
\begin{table}[h]
\centering
\caption{\textbf{Crawl Statistics on Flood-Damage Web Images.} The dataset is highly noisy in long-tail web scenarios.}
\label{tab:crawl_stats}
\small
\begin{tabular*}{\columnwidth}{@{\extracolsep{\fill}} lc}
\toprule
\textbf{Category}        & \textbf{Proportion} \\
\midrule
Target images            & 10\% \\
Thematically related     & 60\% \\
Irrelevant noise         & 30\% \\
\bottomrule
\end{tabular*}
\end{table}

\subsection{Downstream Evaluation on Flooded-Car Recognition}

To further assess whether our curated data genuinely benefits real downstream applications, we conduct two flooded-car recognition tasks that reflect practical inspection scenarios: (i) multi-class classification of flooded automotive parts, and (ii) binary detection of water-damage traces. Both tasks rely on the part and trace taxonomy introduced in the main paper, and all results are reported using Top-1 accuracy on a held-out test set.

\para{Experimental Setup}
For each task, we train two widely used classification backbones---YOLOv8n-cls\cite{yolov8} and ResNet-18\cite{he2016deep}---under two training conditions: models trained on curated \textit{clean} data, and models trained on raw \textit{dirty} web-collected data that contain mixed-quality and incorrect labels. The goal is to quantify how data curation influences downstream reliability, especially in safety-related settings. All hyperparameters and training protocols are kept identical within each model family to ensure a fair comparison.

\para{Part Classification Results}
Table~\ref{tab:flooded_parts_downstream} shows that both architectures achieve near-saturated performance when trained on the clean dataset (YOLOv8n-cls: 0.9729, ResNet-18: 0.9884). However, accuracy drops noticeably when models are trained on the raw noisy dataset (YOLOv8n-cls: 0.8992, ResNet-18: 0.8798). The 8--11 percentage point degradation indicates that even moderately coarse-grained part-level recognition is sensitive to label noise, and highlights the necessity of data refinement.

\begin{table}[t]
\centering
\caption{\textbf{Downstream flooded-car part classification (Top-1 Accuracy).} 
We compare models trained on curated (“clean”) vs.\ raw noisy (“dirty”) web data.}
\label{tab:flooded_parts_downstream}
\begin{tabular}{lcc}
\toprule
\textbf{Model} & \textbf{Training set} & \textbf{Top-1 Acc} \\
\midrule
YOLOv8n-cls  & Clean & 0.9729 \\
YOLOv8n-cls  & Dirty & 0.8992 \\
ResNet-18    & Clean & 0.9884 \\
ResNet-18    & Dirty & 0.8798 \\
\bottomrule
\end{tabular}
\end{table}

\para{Trace Binary Classification Results}
The effect of noise becomes significantly more severe for fine-grained trace detection (Table~\ref{tab:flooded_traces_downstream}). When trained on clean data, both YOLOv8n-cls and ResNet-18 achieve strong performance (0.9519 and 0.9444 respectively). In contrast, training on dirty data collapses performance to near-random levels (0.5667 and 0.5148), demonstrating that noisy supervision fails to provide a meaningful decision boundary for subtle trace cues. This dramatic gap emphasizes that downstream safety-critical tasks require high-quality annotations; noisy web data alone is insufficient.
\begin{table}[t]
\centering
\caption{\textbf{Downstream flooded-car trace binary classification (Top-1 Accuracy).} 
Tasks trained on curated (“clean”) vs.\ raw noisy (“dirty”) data.}
\label{tab:flooded_traces_downstream}
\begin{tabular}{lcc}
\toprule
\textbf{Model} & \textbf{Training set} & \textbf{Top-1 Acc} \\
\midrule
YOLOv8n-cls  & Clean & 0.9519 \\
YOLOv8n-cls  & Dirty & 0.5667 \\
ResNet-18    & Clean & 0.9444 \\
ResNet-18    & Dirty & 0.5148 \\
\bottomrule
\end{tabular}
\end{table}

\para{Summary}
Across both architectures and both tasks, curated clean data consistently yields large and reliable gains. Particularly in the trace classification setting, curation transforms an almost unusable model into a highly reliable one. These results validate that our data-cleaning pipeline not only improves dataset quality metrics but also produces substantial, tangible benefits for downstream flooded-car inspection models.

\section{Psychedelic Mushroom Study Details}
\subsection{Task descriptions}
\begin{figure}[h]
    \centering
    \includegraphics[width=\linewidth]{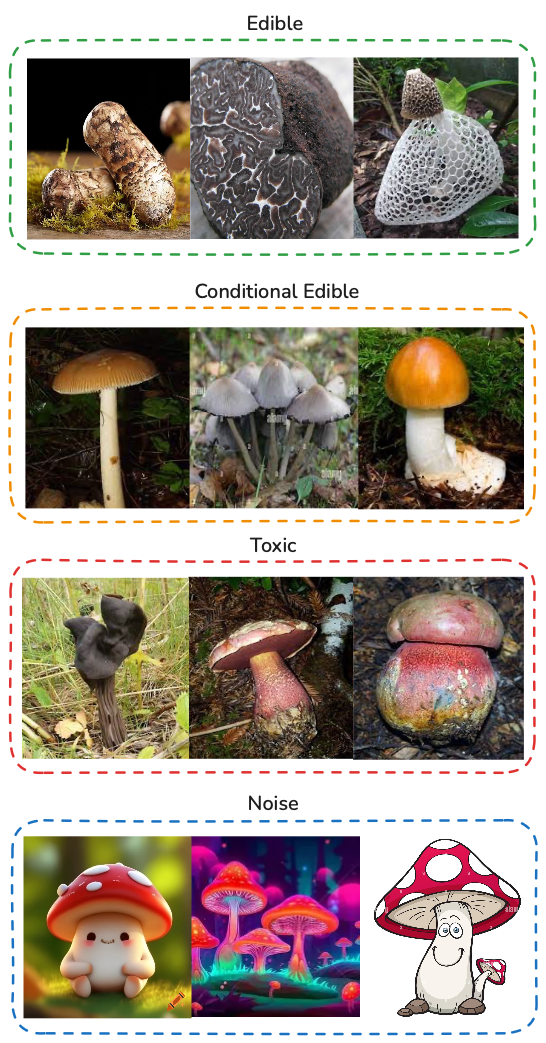}
    \caption{\textbf{Examples of mushroom classes.}
Representative samples from the four categories: conditionally edible, deadly, edible, and noise/stylized. The visual similarity between real and stylized images illustrates the challenge of this task.}
    \label{fig:mush-task}
\end{figure}

In this experiment, both color and shape attributes are defined within closed label sets to ensure consistency and comparability. The color set includes white, red, yellow, brown, and black, representing the primary visual tones observed across mushroom species. The shape set covers strip-shaped, umbrella-shaped, spherical, irregular, and lamellar, corresponding to the most common morphological patterns of mushroom caps and stems. All predictions are constrained to these categories, and synonymous or variant terms (e.g., whitish → white, cap → umbrella-shaped) are automatically normalized during evaluation to maintain label uniformity. Using ReCCur, we further partition the raw web-crawled data into these semantic categories and assign explicit toxicity and attribute labels. As illustrated in Figure~\ref{fig:mush-task}, the mushroom dataset contains both realistic and stylized samples across all four categories, making the task visually ambiguous and prone to noise-induced confusion.

We performed two rounds of large-scale data filtering. Since mushroom images exhibit relatively subtle semantic differences and the web contains a substantial amount of valid samples, only a small number of filtering rounds are required to obtain a dataset with minimal noise. 

\begin{figure}[h]
    \centering
    \includegraphics[width=\linewidth]{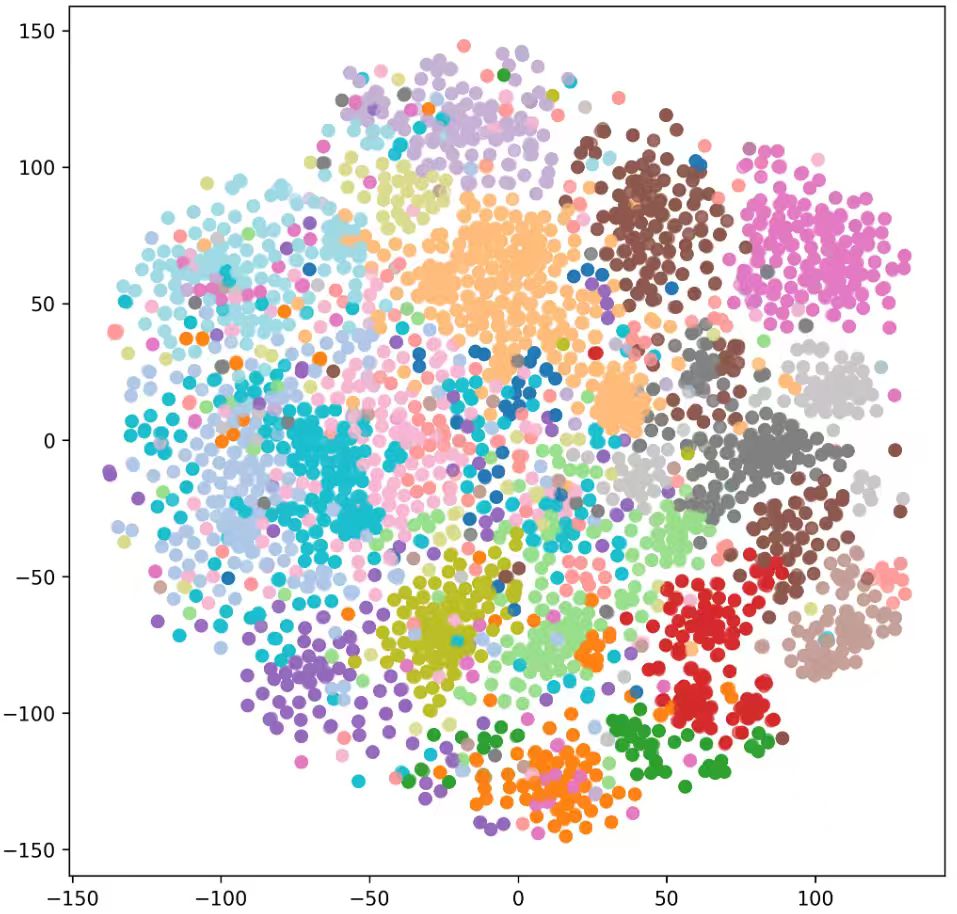}
    \caption{\textbf{Clustered examples of the mushroom dataset after two rounds of filtering.}}
    \label{fig:mushcluster}
\end{figure}

\subsection{Detailed Results}
After the two-stage filtering, we applied clustering to the refined dataset, and the visualization of the clustered images is shown in Figure~\ref{fig:mushcluster}.

\begin{figure}[h]
    \centering
    \includegraphics[width=\linewidth]{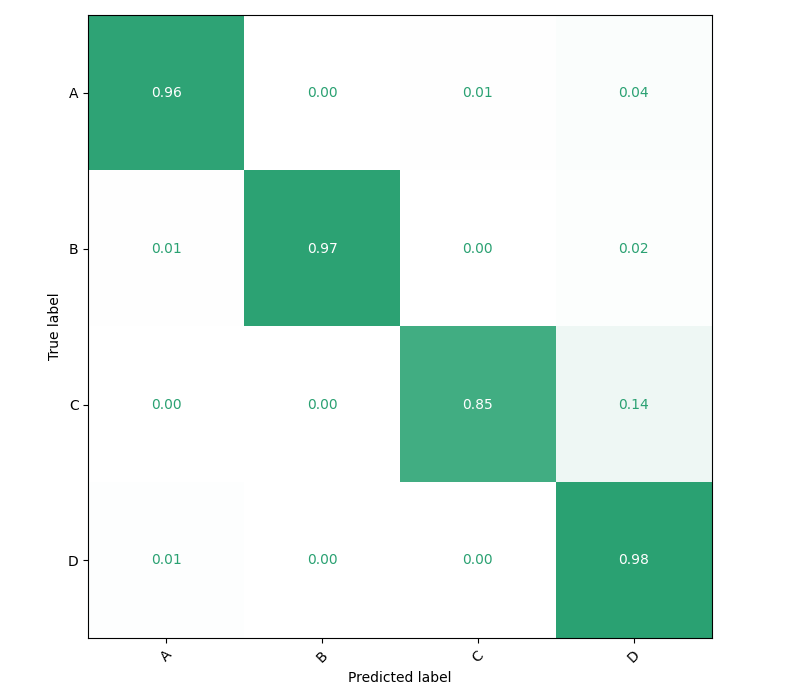}
    \caption{\textbf{Normalized confusion matrix for mushroom classification.}
Most samples lie on the diagonal, with minor confusion between classes C and D. (A: conditionally edible, B: toxic, C: edible, D: noise)}
    \label{fig:mushroom_confmat}
\end{figure}

As shown in Figure~\ref{fig:mushroom_confmat}, a small portion of edible samples were misclassified as noise, slightly reducing the data retention rate. Table~\ref{tab:mushroom_table} demonstrates the details of each group.

\begin{table}[h]
\centering
\caption{\textbf{Classification results for mushroom dataset}}
\label{tab:mushroom_table}
\small
\setlength{\tabcolsep}{8pt}
\renewcommand{\arraystretch}{1.15}
\begin{tabular}{lccc}
\toprule
\textbf{Class} & \textbf{Precision} & \textbf{Recall} & \textbf{F1-score} \\
\midrule
A & 0.9786 & 0.9567 & 0.9675 \\
B & 0.9961 & 0.9716 & 0.9837 \\
C & 0.9926 & 0.8522 & 0.9170 \\
D & 0.8209 & 0.9842 & 0.8951 \\
\midrule

Macro Avg  & 0.9470 & 0.9412 & 0.9409 \\

\bottomrule
\end{tabular}
\end{table}

\subsection{Downstream Evaluation on Psychedelic Mushroom Classification}

The psychedelic mushroom domain represents a particularly challenging setting for data curation and downstream recognition. Unlike the flooded-car or wall-damage datasets, mushroom images collected from the web exhibit large stylistic variability: real photographs, field-guide scans, high-saturation illustrations, AI-generated artworks, and meme-like ``psychedelic’’ images frequently coexist. Many of these images share similar global color and texture patterns despite belonging to fundamentally different semantic categories (real vs.\ fake), making this domain an ideal testbed for evaluating whether ReCCur can mitigate noise-induced ambiguity and support reliable downstream classification.

\para{Task motivation}
In practice, mushroom-related image collections are often polluted by synthetic or stylized content, including cartoons, fantasy illustrations, and highly edited images that do not reflect realistic morphology. For both scientific analysis and dataset construction, it is important to distinguish real mushroom photographs from non-photographic or fake imagery. At the same time, we are also interested in whether cleaned data help models retain fine-grained distinctions between different mushroom categories when such fake content is present in the training pool. Therefore, we evaluate two types of downstream tasks to measure the practical effect of ReCCur-curated data.

\para{Experimental Setup}
We consider two complementary downstream tasks:
\begin{itemize}
    \item \textbf{Multi-class recognition:} four-class classification among \textit{conditionally edible}, \textit{deadly}, \textit{edible}, and \textit{noise/stylized} samples. 
    \item \textbf{Real-vs-fake binary recognition:} binary classification between \textit{real} mushroom photographs and \textit{fake/non-photographic} images (including cartoons, AI-generated images, and heavily stylized or meme-like content). This reflects the ability to separate realistic biological imagery from synthetic or illustrative content.
\end{itemize}

For both tasks, we train two representative classification backbones—YOLOv8n-cls\cite{yolov8} and ResNet-18\cite{he2016deep}—under two data regimes: (i) our ReCCur-curated \textit{clean} dataset, and (ii) the original \textit{dirty} web-crawled dataset, which contains mislabeled samples and a large proportion of stylized or fake imagery. Hyperparameters, data augmentations, and training schedules are kept identical within each architecture to ensure fair comparisons. All metrics are reported as Top-1 accuracy on a held-out clean test split.

\vspace{0.5em}
\para{Multi-class recognition results}
Table~\ref{tab:mushroom_multicls_downstream} reports the Top-1 accuracy for the four-class setting. Training on clean data substantially improves model reliability: YOLOv8n-cls improves from 0.852 (dirty) to 0.948 (clean), and ResNet-18 improves from 0.875 to 0.964. These gains show that raw web data still introduce non-trivial inter-class confusion—particularly between edible, conditionally edible, and stylized psychedelic images whose color and shape patterns resemble real species. By removing ambiguous, off-topic, and strongly stylized samples, ReCCur allows both architectures to form sharper decision boundaries and reduces systematic mislabeling caused by style bias.

\begin{table}[t]
\centering
\caption{\textbf{Downstream mushroom multi-class classification (Top-1 Accuracy).}}
\label{tab:mushroom_multicls_downstream}
\begin{tabular}{lcc}
\toprule
\textbf{Model} & \textbf{Training set} & \textbf{Top-1 Acc} \\
\midrule
YOLOv8n-cls  & Clean & 0.948 \\
YOLOv8n-cls  & Dirty & 0.852 \\
ResNet-18    & Clean & 0.964 \\
ResNet-18    & Dirty & 0.875 \\
\bottomrule
\end{tabular}
\end{table}

\vspace{0.5em}
\para{Real-vs-fake binary results}
Noise has a clear impact on the real-vs-fake classification task, where the model must distinguish realistic biological imagery from cartoons, AI-generated art, and heavily edited “psychedelic’’ content. As shown in Table~\ref{tab:mushroom_binary_downstream}, training on clean data yields very strong performance, with Top-1 accuracy of 0.972 for YOLOv8n-cls and 0.981 for ResNet-18. When trained on the dirty dataset, the same architectures remain highly accurate (0.914 and 0.923, respectively), but still fall short of their clean counterparts by roughly 6 percentage points. This consistent drop suggests that mixed, unlabeled fake content in the training set perturbs the decision boundary between real and non-real imagery, causing more confusions between stylized mushrooms and genuine specimens. ReCCur’s curated dataset alleviates this problem by filtering out non-photographic or semantically inconsistent samples before training.

\begin{table}[t]
\centering
\caption{\textbf{Downstream mushroom real-vs-fake classification (Top-1 Accuracy).}}
\label{tab:mushroom_binary_downstream}
\begin{tabular}{lcc}
\toprule
\textbf{Model} & \textbf{Training set} & \textbf{Top-1 Acc} \\
\midrule
YOLOv8n-cls  & Clean & 0.972 \\
YOLOv8n-cls  & Dirty & 0.914 \\
ResNet-18    & Clean & 0.981 \\
ResNet-18    & Dirty & 0.923 \\
\bottomrule
\end{tabular}
\end{table}

\vspace{0.5em}
\para{Summary}
Across both multi-class and real-vs-fake tasks, curated data deliver consistent and substantial improvements for both YOLOv8n-cls and ResNet-18. In the real-vs-fake setting, ReCCur further boosts already strong models into a near-saturated regime, cutting error rates by more than half compared with training on dirty data. These findings reinforce that structured multimodal filtering is crucial in domains with heavy stylistic noise and synthetic content, and that the benefits of curation propagate directly into downstream mushroom-related recognition tasks.

\section{Case Study: Wall-Damage Detection}
\vspace{1em}
\para{Background} Wall aging is a rare, under-represented corner case in scene understanding and defect detection, with diverse manifestations (stains, efflorescence, peeling, micro-cracks). Its scarcity and label noise induce distribution shift—models trained on pristine surfaces overfit texture/color priors—so robustness to aging artifacts remains underexplored and necessary.

\begin{table*}[t]
\centering
\caption{\textbf{Method Comparison for the Wall-Damage study.} GPT-based methods vs. ReCCur.}
\label{tab:thirdcase}
\begingroup
\footnotesize
\setlength{\tabcolsep}{6pt}            
\renewcommand{\arraystretch}{1.15}     
\newcommand{\NA}{\textemdash}          
\begin{tabular*}{\textwidth}{@{\extracolsep{\fill}} lcccccccccc @{}}
\toprule
\textbf{Method} &
\multicolumn{5}{c}{\textbf{Image Recognition and Noise Filtering Stage}} &
\multicolumn{4}{c}{\textbf{Semantic Labeling Stage}} \\
\cmidrule(lr){2-6}\cmidrule(lr){7-10}
& \textbf{Precision} & \textbf{Recall} & \textbf{F1 score} & \textbf{NRR} & \textbf{CDRR}
& \textbf{Precision} & \textbf{Recall} & \textbf{F1 score} & \textbf{Perfect match} \\
\midrule
GPT\mbox{-}5                            & 0.889 & 0.865 & 0.883 & 0.824 & 0.901 & 0.814 & 0.869 & 0.822 & 0.772 \\
Retrieval\mbox{-}Augmented GPT\mbox{-}5 & 0.921 & 0.903 & 0.896 & 0.897 & 0.924 & 0.817 & 0.861 & 0.828 & 0.786 \\
ReCCur (ours)                           & 0.968 & 0.965 & 0.937 & 0.965 & 0.984 & 0.862 & 0.887 & 0.863 & 0.854 \\
\bottomrule
\end{tabular*}
\endgroup
\vspace{-0.4em}
\end{table*}

\vspace{1em}
\para{Implementation Details} We build on the S2DS wall dataset \cite{BenzRodehorst_GCPR_2022_S2DS} and apply ReCCur for data expansion, generating four additional wall material domains: brick, granite, metal, and wood, while also annotating noise samples. After three rounds of recognition–filtering, we obtain the final dataset. All settings are kept identical to those used in the flood-damaged vehicle experiments.

\vspace{1em}
\para{Result} In this scenario, ReCCur performs strongly in both data classification and semantic labeling, consistently surpassing the GPT-based baselines. Increasing the number of refinement rounds further improves accuracy, confirming the effectiveness of ReCCur. The normalized confusion matrix shows dominant diagonal mass with only minor off-diagonal leakage, indicating well-separated classes.
A cross-method comparison further demonstrates ReCCur’s advantage over baselines across precision/recall/F1 (Table~\ref{tab:thirdcase}).
The results show that all methods perform better in detecting large-scale traces than in identifying fine or small ones. ReCCur, benefiting from its two-stage reasoning design, achieves slightly superior overall performance compared with other approaches.
\begin{figure}[h]
    \centering
    \includegraphics[width=\linewidth]{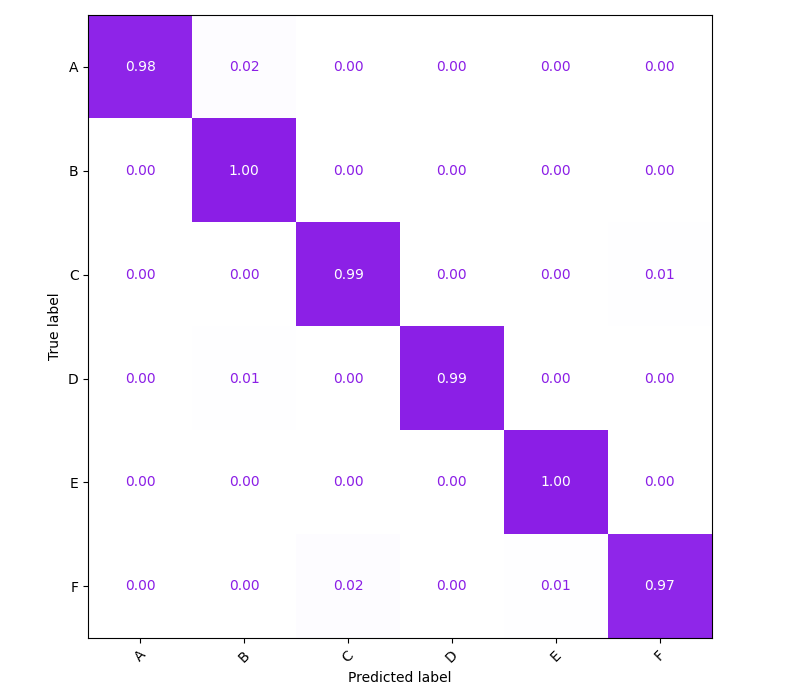}
    \caption{\textbf{Normalized confusion matrix for wall classification.}
    A: brick, B: granite, C: metal, D: concrete, E: noise, F: wood.}
    \label{fig:wall_confmat}
\end{figure}

\section{Limitations and Future Work}
While ReCCur demonstrates that a training-free-core, multi-stage curation pipeline can substantially improve the purity and semantic richness of noisy web data under modest compute, several limitations remain.

\paragraph{Scope of domains and modalities.}
Our empirical validation focuses on three case studies (flooded-vehicle inspection, psychedelic mushrooms, and wall-damage detection), all in the image domain. Although these settings already exhibit significant long-tail noise and semantic ambiguity, they do not cover video, 3D, or temporal corner cases, nor highly interactive embodied scenarios. Extending ReCCur to multi-frame evidence, depth/3D representations, and sequential phenomena (e.g., progressive damage, time-varying scenes) is an important next step.

\paragraph{Dependence on upstream models.}
The pipeline relies on pre-trained VLMs and visual encoders (e.g., ChatGPT-5, Qwen-VL 2.5, CLIP, DINOv2, BEiT) for keyword expansion, multimodal similarity scoring, and semantic reasoning. As a result, its failure modes inherit biases and blind spots from these models, and performance may degrade on domains that are poorly represented in their pre-training corpora. Future work includes stress-testing ReCCur with stronger and weaker backbones, incorporating lightweight domain adaptation, and explicitly modeling epistemic uncertainty from upstream models.

\paragraph{Residual noise and coverage gaps.}
Although ReCCur achieves high precision and strong clean-data retention on our benchmarks, it cannot guarantee perfectly noise-free labels or exhaustive coverage of all rare phenomena. In particular, trace-level patterns (e.g., subtle dust/sand, fine-grained texture artifacts) remain challenging, and the current thresholds for Confidence Activation and uncertainty routing are tuned heuristically. A promising direction is to couple ReCCur with formal calibration objectives and probabilistic guarantees on curation quality, as well as adaptive thresholding driven by downstream performance.
\paragraph{Human supervision and evaluation design.}
The human-in-the-loop components are intentionally lightweight, but they still require annotators with domain knowledge (e.g., automotive experts) and may not be universally available. Moreover, while we do provide downstream evaluations to verify the practical value of curated data, the scope of our analysis remains limited. We have not yet systematically compared ReCCur-curated data against a broader range of alternative filtering baselines or large-scale industrial systems, and our downstream tasks cover only a subset of possible real-world applications. Future work includes integrating ReCCur into end-to-end training pipelines, measuring downstream impact on safety-critical tasks, and exploring interfaces that allow non-experts to contribute reliable feedback.

\paragraph{Systemic and distributional biases.}
Because our web-crawled data reflect existing online content, they may inherit regional, socio-economic, or aesthetic biases (e.g., overrepresentation of certain car models, building materials, or mushroom species). ReCCur currently improves label quality but does not explicitly correct such imbalances. An important extension is to integrate fairness- and diversity-aware objectives into both acquisition (keyword expansion, crawling) and filtering (sampling, cluster selection), and to provide users with tools for auditing and re-balancing curated datasets.

\section{Ethics and Societal Impact}
ReCCur is designed as a general-purpose, low-compute curation framework for corner cases, with the goal of improving robustness and transparency in safety-relevant applications. Our case studies focus on domains with clear consumer- or public-safety motivations: flooded-vehicle inspection, toxic mushroom identification, and wall-damage detection. In these settings, higher-quality corner-case datasets can help reduce safety risks (e.g., resale of flood-damaged vehicles, accidental poisoning, or structural failures) by supporting better detection models and more reliable diagnostic tools.

At the same time, the ability to systematically mine, filter, and semantically enrich rare visual patterns carries dual-use risks. In principle, similar pipelines could be repurposed to curate sensitive or privacy-relevant data, or to build datasets that facilitate pervasive surveillance or targeted profiling. Our design choices partially mitigate these risks by (i) restricting experiments to clearly task-focused domains, (ii) emphasizing transparent, evidence-based labeling rather than hidden user profiling, and (iii) using human supervision primarily to validate technical labels (e.g., parts, traces, material types) rather than personal attributes.

All data used in our experiments are sourced from publicly accessible web content or existing datasets, and we avoid crawling obvious private or access-controlled sources. We do not attempt to infer demographic or sensitive attributes of individuals, and we do not release any identity annotations. Any future public release of curated datasets will exclude images that predominantly depict identifiable individuals, and we will make best efforts to blur or mask residual personally identifiable information (such as faces or license plates) when such content is incidentally present. We explicitly discourage using ReCCur or its curated datasets for applications that violate local regulations on privacy, surveillance, or data protection.

Finally, our work should be viewed as early-stage research on data curation rather than a turnkey deployed system. Any real-world deployment—especially in regulated, high-stakes domains such as insurance, consumer protection, infrastructure inspection, or healthcare—must undergo separate risk assessment, domain-specific validation, and legal/ethical review. Practitioners remain responsible for ensuring that their use of ReCCur and related datasets complies with applicable laws, institutional review standards, and societal norms.

\section{License and Legal Compliance}
Upon acceptance, we will publicly release both the ReCCur code and curated datasets for research and commercial use, subject to clear licensing and compliance safeguards. The full training, evaluation, and inference code will be hosted on GitHub under a permissive open-source license (e.g., Apache 2.0), allowing users to use, modify, and integrate ReCCur into academic projects and commercial products.

The curated ReCCur datasets will be released under a dedicated “ReCCur Data License” that allows both non-commercial and commercial use, modification, and redistribution, provided that users (i) give appropriate credit to our work, (ii) redistribute any derived datasets under the same data license terms, and (iii) comply with all applicable copyright, privacy, and data-protection laws in their jurisdictions. The license will explicitly prohibit use of the data for biometric identification, mass surveillance, or any other applications that reasonably threaten individual privacy or civil liberties.

Our curation pipeline builds on third-party models and datasets (e.g., CLIP, DINOv2, BEiT, VLM APIs, and existing defect datasets) and may include web-crawled images. For each upstream resource, we will document its source, version, and original license in a “Licenses and Attributions” section. We will only redistribute third-party content when their licenses permit this; otherwise, we will provide download links and preprocessing scripts so that users can recreate our datasets themselves. For web-crawled content, we treat the curated data as derivative research artifacts, store provenance metadata where possible, and provide a clear takedown channel: verified requests from original content owners will lead to removal of the corresponding samples in subsequent releases.

Any deployment of ReCCur or its derivatives in real-world decision-making pipelines must comply with relevant legal frameworks (e.g., data protection, consumer protection, product liability, and sector-specific regulations). Our release will therefore include a disclaimer that the authors and institutions assume no responsibility for misuse or unlawful applications of the code or datasets, and that downstream users bear full responsibility for ensuring regulatory compliance in their own jurisdictions.

\begin{table*}[h]
\centering
\footnotesize
\caption{\textbf{List of symbols used in this work.}}
\resizebox{\textwidth}{!}{
\begin{tabular}{llll}
\hline
\textbf{Symbol} & \textbf{Type / Space} & \textbf{Meaning} \\
\hline
\textbf{Datasets / Sets} \\
$\mathcal{D}$ & $\mathcal{P}(\mathbb{R}^{H\times W\times 3})$ & Original large-scale web-crawled dataset \\
$\mathcal{D}_c$ & $\mathcal{P}(\mathbb{R}^{H\times W\times 3})$ & Dataset of category $c$ \\
$\mathcal{D}_{\mathrm{tri}}$ & $\mathcal{P}\big(\mathbb{R}^{H\times W\times 3} \times \Sigma^\ast \times \Sigma^\ast\big)$ & Trimodal sample triplet set $\{\mathbf{u}_i\}_{i=1}^N$ \\
$\Pi$ & $\mathcal{P}(\mathbb{R}^{H\times W\times 3})$ & Refined dataset after multimodal filtering \\
$\mathcal{C}=\{\mathcal{C}_1,\dots,\mathcal{C}_j\}$ & $\mathcal{P}\big(\mathcal{P}(\mathbb{R}^{1536})\big)$ & Clusters obtained by grouping embeddings \\
$\mathcal{A}=\{(X_i,y_i)\}$ & $\mathcal{P}\big(\mathbb{R}^{H\times W\times 3}\times\mathcal{Y}\big)$ & Manually annotated reference dataset \\
$\mathcal{I}$ & $(\mathcal{P}(\mathcal{Y}\times\mathbb{R}^{d_{768}}))\times(\mathcal{P}(\mathcal{Y}\times\mathbb{R}^{d_{1024}}))\times(\mathcal{P}(\mathcal{Y}\times\mathbb{R}^{d_{1024}}))$ & Vector index database of three expert sub-databases \\
$\pi$ & $\mathcal{P}(\mathbb{R}^{H\times W\times 3})$ & Final integrated dataset after iterations \\
$\mathcal{I}_{\mathrm{CLIP}}$ & $\mathcal{P}\big(\mathcal{Y}\times\mathbb{R}^{d_{768}}\big)$ & Sub-database of CLIP (label–vector pairs) \\
$\mathcal{I}_{\mathrm{DINOv2}}$ & $\mathcal{P}\big(\mathcal{Y}\times\mathbb{R}^{d_{1024}}\big)$ & Sub-database of DINOv2 (label–vector pairs) \\
$\mathcal{I}_{\mathrm{BEiT}}$ & $\mathcal{P}\big(\mathcal{Y}\times\mathbb{R}^{d_{1024}}\big)$ & Sub-database of BEiT (label–vector pairs) \\
\hline
\textbf{Samples / Variables} \\
$X_i$ & $\mathbb{R}^{H\times W\times 3}$ & $i$-th image tensor \\
$H_i$ & $\Sigma^\ast$ & VLM-generated textual description of $X_i$ \\
$Q_i$ & $\Sigma^\ast$ & Search query keywords \\
$\mathbf{u}_i=(X_i,H_i,Q_i)$ & $\mathbb{R}^{H\times W\times 3}\times \Sigma^\ast \times \Sigma^\ast$ & Trimodal sample triplet \\
$\mathbf{z}_i$ & $\mathbb{R}^{1536}$ & Semantically enhanced embedding (CLIP-img $\Vert$ CLIP-txt) \\
$r_i$ & $\{0,1\}$ & Binary correlation label (1=high, 0=low) \\
$\bar{r}_j$ & $[0,1]$ & Average cluster relevance score \\
$y_i$ & $\mathcal{Y}$ & Ground-truth class label \\
$\hat{y},\,\hat{y}^{(m)}$ & $\mathcal{Y}$ & Predicted label (ensemble / expert $m$) \\
$\hat{y}_{\text{sem-1}},\,\hat{y}_{\text{sem-2}},\,\hat{y}_{\text{sem}}$ & $\mathcal{Y}\times\mathcal{P}(\boldsymbol{\Phi})$ & Semantic labels at reasoning stages \\
$\mathcal{Y}$ & - & Augmented label space used by $\mathrm{Dec}(X)$ \\
\hline
\textbf{Prompts / Keywords} \\
$\mathcal{P}_{\mathrm{init}},\,\mathcal{P}_{\mathrm{opt}},\,\mathcal{P}_c$ & $\Sigma^\ast$ & Initial / optimized / category prompts \\
$\mathcal{K}_{\mathrm{V}},\,\mathcal{K}_{\mathrm{T}},\,\mathcal{K}$ & $\Sigma^\ast$ & VLM keywords / text-only keywords / final corpus \\
$\mathcal{G}_{\mathrm{V}}(\cdot)$ & $\mathcal{P}(\mathbb{R}^{H\times W\times 3}) \times \Sigma^\ast \to \Sigma^\ast$ & VLM-based keyword generation \\
$\mathcal{G}_{\mathrm{T}}(\cdot)$ & $\Sigma^\ast \to \Sigma^\ast$ & Text-only keyword generation \\
$\mathcal{L}(\cdot)$ & $\Sigma^\ast \to \Sigma^\ast$ & Multilingual lexicon expansion \\

\textbf{Models / Functions} \\
$\mathcal{E}_{\mathrm{CLIP}},\,\mathcal{E}_{\mathrm{DINOv2}},\,\mathcal{E}_{\mathrm{BEiT}}$ & $\mathbb{R}^{H\times W\times 3}\to \mathbb{R}^{d_m}$; $\Sigma^\ast\to \mathbb{R}^{d_m}$ & Expert image/text encoders \\
$\mathcal{F}(\cdot)$ & $\mathbb{R}^3\to \mathbb{R}$ & Multimodal similarity fusion operator \\
$\mathrm{Sim}_{\mathrm{mm}}(\mathbf{u}_i)$ & $\mathbb{R}$ & Multimodal similarity score \\
$\mathsf{Prop}(X,\mathcal{P}_c)$ & $\mathbb{R}^{H\times W\times 3}\times \Sigma^\ast \to (\mathbb{Z}^{4})^{N_g}\times \{0,1\}^{N_g}$ & VLM Proposer (region extraction and feature flags) \\
$\psi(\mathbf{r}_{c,g}^{(j)},\boldsymbol{\Phi})$ & $\{0,1\}$ & Indicator for region–attribute presence \\

$\operatorname{Dec}(X)$ & $\mathbb{R}^{H\times W\times 3}\to (\mathcal{Y}\cup\{\text{non-target}\})$ & Final decision rule based on confidence thresholds \\
\hline
\textbf{Experts / Embeddings / Similarity} \\
$\mathcal{M}$ & $\{\mathrm{CLIP},\mathrm{DINOv2},\mathrm{BEiT}\}$ & Expert ensemble \\
$\mathcal{N}^{(m)}=\big((\xi^{(m)}_k,\,y^{(m)}_k)\big)_{k=1}^{K}$ & $\big([-1,1]\times \mathcal{Y}\big)^{K}$ & Top-$K$ neighbor similarity–label sequence (sorted by $\xi$) \\
$\xi_k$ & $[-1,1]$ & Cosine similarity value \\

$\overline{\mathrm{FAS}}(X,c_k),\ \overline{\mathrm{FAS}}(X,\hat{y})$ & $(\mathbb{R}^{H\times W\times 3}\times \mathcal{Y})\to[-1,1]$ & Feature Alignment Scores (per-class / averaged) \\
$\mathbf{\mu}_c^{(m)}$ & $\mathbb{R}^{d_m}$ & Mean embedding of class $c$ under expert $m$ \\
\hline
\textbf{Confidence / Thresholds / Scores} \\
$\operatorname{Conf}_{\mathrm{l}}^{(m)}(X)$ & $[-1,1]$ & label confidence for expert $m$ \\
$\mathrm{Conf}_{\mathrm{t}}(X),\,\overline{\mathrm{Conf}_{\mathrm{l}}}(X)$ & $[-1,1]$ & Topic Confidence and Label Confidence \\
$\theta_{\mathrm{t}},\,\theta_{\mathrm{l}}$ & $[-1,1]$ & Confidence thresholds \\
$\tau_{\mathrm{mm}}$ & $\mathbb{R}$ & Threshold for multimodal similarity filtering \\
$B(X)$ & $[-1,1]$ & Boundary strength (maximum alignment) \\
$y^\star(X)$ & $\mathcal{Y}$ & Boundary attribution class \\
\hline
\textbf{Regions / Features} \\
$\mathcal{R}_{c,g}=\{\mathbf{r}_{c,g}^{(1)},\dots,\mathbf{r}_{c,g}^{(N_g)}\}$ & $(\mathbb{Z}^{4})^{N_g}$ & Subregion bounding boxes at granularity $g$ (each as $(x,y,w,h)$ in pixels; typically $N_g=g^2$) \\
$\mathbf{r}_{c,g}^{(j)}$ & $\mathbb{Z}^{4}$ & $j$-th subregion box $(x,y,w,h)$ in pixels \\
$N_g$ & $\mathbb{N}$ & Number of subregions at granularity $g$ (for $g\times g$ grid, $N_g=g^2$) \\
$\boldsymbol{\Phi}$ & -- & Visual attribute set (e.g., rust, mold, mud) \\

$\mathcal{R}_{s,g}=\{\mathbf{r}_{s,g}^{(1)},\dots,\mathbf{r}_{s,g}^{(N_g)}\}$ & $(\mathbb{Z}^{4})^{N_g}$ & Subject subregion set at granularity $g$ ($\mathcal{R}_{s,g}\subset\mathcal{R}_{c,g}$) \\
$\mathbf{r}_{s,g}^{(j)}$ & $\mathbb{Z}^{4}$ & $j$-th subject subregion box $(x,y,w,h)$ \\
\hline
\textbf{Scalars / Parameters} \\
$K,\,K_L,\,K_H$ & $\mathbb{N}$ & \#neighbors / candidate pool sizes \\
$\alpha$ & $(0,1]$ & Sampling ratio for low-alignment pool \\

$N,\,N_c$ & $\mathbb{N}$ & Dataset size / per-class sample count \\
$N_{\mathrm{per}}$ & $\mathbb{N}$ & Number of images manually annotated per cluster \\
$t$ & $\mathbb{N}$ & Number of classes used in derivations (e.g., boundary ranking) \\
$T$ & $\mathbb{R}_{>0}$ & Softmax temperature parameter \\
\hline
\end{tabular}}
\label{tab:SymbolsASD}
\end{table*}

{
    \small
    \bibliographystyle{ieeenat_fullname}
    \bibliography{main}
}

\end{document}